\documentclass[review]{elsarticle}
\usepackage{setspace}
\usepackage{amssymb}
 \usepackage{amsmath} 
\usepackage{multirow}
\usepackage[normalem]{ulem}
\useunder{\uline}{\ul}{}
\usepackage{lineno,hyperref}
\usepackage{algorithm}
\usepackage[margin=1in]{geometry}
\usepackage{algpseudocode}
\usepackage{adjustbox}
\usepackage{graphicx}
\usepackage{lipsum}
\usepackage[table,xcdraw]{xcolor}
\usepackage[justification=centering]{caption}
\usepackage{caption}
\usepackage{multirow}
\usepackage{etoolbox}
\usepackage{subfigure}  
\usepackage[margin=1in]{geometry}
\usepackage{caption}
\usepackage{hyperref}
\usepackage{array} 
\usepackage{array}
\usepackage{subcaption}
\usepackage{rotating} 
\usepackage[margin=1in]{geometry}
\usepackage{ amssymb }
\usepackage{multirow}
\usepackage{subfigure}
\usepackage{comment}
\usepackage{subcaption}
\usepackage{placeins}
\usepackage{bm}
\usepackage{listings}

\usepackage{xcolor}
\usepackage{amssymb}
\renewcommand{\arraystretch}{1.5}
\definecolor{codegreen}{rgb}{0,0,1}
\definecolor{codegray}{rgb}{0.95,0.5,0.5}
\definecolor{codepurple}{rgb}{0.58,0,0.82}
\definecolor{backcolour}{rgb}{1, 1, 1}

\lstdefinestyle{mystyle}{
    backgroundcolor=\color{backcolour},   
    commentstyle=\color{codegreen},
    keywordstyle=\color{magenta},
    numberstyle=\tiny\color{codegray},
    stringstyle=\color{codepurple},
    basicstyle=\large\ttfamily\footnotesize,
    breakatwhitespace=false,         
    breaklines=true,                 
    captionpos=b,                    
    keepspaces=true,                 
    numbersep=5pt,                  
    showspaces=false,                
    showstringspaces=false,
    showtabs=false,                  
    tabsize=2
}
\usepackage{amssymb}

\journal{arXiv}

\begin{document}

\begin{frontmatter}



\title{{\color{black}DB2-TransF: All You Need Is Learnable Daubechies Wavelets for Time Series Forecasting }}


\author[inst1]{Moulik Gupta, Achyut Mani Tripathi}

\affiliation[inst1]{organization={Department of Computer Science \& Engineering \& Communication \& Engineering},
            addressline={G B Pant}, 
            city={New Delhi},
            postcode={580007}, 
            state={Delhi},
            country={India}}
\affiliation[inst2]{organization={Department of Computer Science \& Engineering},
            addressline={Indian Institute of Technology}, 
            city={Dharwad},
            postcode={580007}, 
            state={Karnataka},
            country={India}}

\begin{abstract}
Time series forecasting requires models that can efficiently capture complex temporal dependencies, especially in large-scale and high-dimensional settings. While Transformer-based architectures excel at modeling long-range dependencies, their quadratic computational complexity poses limitations on scalability and adaptability. To overcome these challenges, we introduce DB2-TransF, a novel Transformer-inspired architecture that replaces the self-attention mechanism with a learnable Daubechies wavelet coefficient layer. This wavelet-based module efficiently captures multi-scale local and global patterns and enhances the modeling of correlations across multiple time series for the time series forecasting task. Extensive experiments on 13 standard forecasting benchmarks demonstrate that DB2-TransF achieves comparable or superior predictive accuracy to conventional Transformers, while substantially reducing memory usage for the time series forecasting task. The obtained experimental results position DB2-TransF as a scalable and resource-efficient framework for advanced time series forecasting. Our code is available at \href{red}{https://github.com/SteadySurfdom/DB2-TransF}
\end{abstract}


\begin{highlights}
 \item We propose DB2-TransF, a linear time series forecasting model that incorporates a novel learnable Daubechies Wavelet module as an alternative to the standard self-attention mechanism, enhancing both predictive accuracy and computational efficiency.

\item  The DB2-TransF model demonstrated consistent performance gains over leading forecasting methods, while also reducing computational requirements compared to earlier transformer and MLP-based models.

\item  We conducted detailed empirical analysis to uncover the model’s behavior across 13 diverse time series forecasting datasets, highlighting its effectiveness and generalization capability in time series forecasting tasks.

\end{highlights}
\begin{keyword}
Daubechies Wavelets, Deep Learning, Forecasting, Multivariate -Time Series, Transformer. 
\end{keyword}
\end{frontmatter}
\section{Introduction
}
Time series data are defined as logs of observations collected over time, often featuring a lot of volume, high dimensionality, and constant change. These types of data are usually classified into univariate and multivariate time series. In the modern data-centric society, the occurrence of multivariate time series (MTS) is on the rise since it captures several variables over a certain period.  Multivariate time series forecasting (TSF) has received significant attention due to its wide range of applications, including energy consumption prediction \cite{khan2020towards}, meteorological forecasting \cite{angryk2020multivariate}, and traffic control \cite{chen2001freeway}. One of the major challenges in this domain is effectively modeling the correlations among multiple variables, which becomes increasingly difficult when using traditional machine learning or statistical approaches. Classical models such as Vector Autoregression (VAR) \cite{akaike1969fitting} and Autoregressive Integrated Moving Average (ARIMA) \cite{min2015wind} have been widely employed across various TSF tasks. While these statistical methods perform reasonably well, they suffer from key limitations: reliance on handcrafted features, difficulty in capturing non-linear patterns, and scalability issues as the number of time series increases. Machine learning techniques offer a compelling alternative by enabling automatic extraction of meaningful representations directly from raw time series data. The rapid progress in deep learning has further inspired the development of more accurate TSF models, including multilayer perceptrons (MLPs) \cite{khashei2010artificial}, recurrent neural networks (RNNs) \cite{assaad2008new, zhang2023robust}, convolutional and temporal convolutional networks (CNNs/TCNs) \cite{semenoglou2023image,dudukcu2023temporal}, and Transformer-based architectures \cite{liu2023itransformer,zhou2021informer}. They excel at capturing temporal dependencies as well as modeling complex non-linear relationships among several interacting factors while enabling end-to-end learning that minimizes manual feature construction efforts and streamlines the forecasting process. A distinguishing characteristic of multivariate time series (MTS) data, unlike other sequential data types, is that only a single scalar value is recorded at each timestamp. As a result, individual observations tend to be context-poor, semantically rich, but very noisy. To improve the understanding of such data, a number of deep learning techniques \cite{assaad2008new, zhang2023robust, semenoglou2023image,dudukcu2023temporal, liu2023itransformer, zhou2021informer} have been developed that make use of historical sequences to learn temporal patterns and improve the model's proficiency in learning meaningful representations. Whereas these approaches focus on advancing models for the extraction of temporal trends, real-world MTS data pose some distinctive problems owing to their complexity and heterogeneity. Additionally, multiple time series data also suffer from redundancy problems. In practical scenarios like monitoring traffic, sensors (like cameras or inductive loops) tend to collect data very frequently (for example, every second) in order to provide continuous situational awareness. However, this high-frequency sampling does not always provide useful information because most readings would be redundant. The challenges associated with time series forecasting (TSF) are further amplified when deploying deep learning models, particularly transformer-based architectures, in resource-constrained environments. This necessitates the design of deep models that not only maintain high predictive accuracy under limited computational resources but also address the quadratic complexity inherent to transformer models. In response to this, we propose a novel deep architecture based on learnable Daubechies Wavelets, which effectively captures the temporal dynamics and inter-variable correlations within multivariate time series. Moreover, the proposed multiscale learnable DB2 block offers an efficient alternative to the conventional self-attention mechanism \cite{vaswani2017attention} by effectively capturing both noise components and smooth temporal trends in the time series, while significantly reducing computational overhead. Our model demonstrates superior performance and computational efficiency compared to existing state-of-the-art transformer-based approaches \cite{liu2023itransformer,zhou2021informer}. \par

The significant contributions of our paper are as follows:

\begin{itemize}
   \item We propose DB2-TransF, a linear time series forecasting model that incorporates a novel learnable Daubechies Wavelet module as an alternative to the standard self-attention mechanism, enhancing both predictive accuracy and computational efficiency.

\item  The DB2-TransF model demonstrated consistent performance gains over leading forecasting methods, while also reducing computational requirements compared to earlier transformer and MLP-based models.

\item  We conducted a detailed empirical analysis to uncover the model’s behavior across 13 diverse time series forecasting datasets, highlighting its effectiveness and generalization capability in time series forecasting tasks.
\end{itemize}

\section{Related Work}
The various techniques proposed for time series forecasting (TSF) can be broadly categorized into three groups: RNN-based models, CNN-based models, and Transformer-based models. For a clearer comparison, we have grouped and compared the different models within each category in Table \ref{LS_RNN}, Table \ref{LS_CNN}, and Table \ref{LS_TSF}, respectively. Table \ref{LS_RNN} summarizes various RNN-based methods proposed for the TSF task. Although these models have shown strong capabilities in modeling temporal dependencies, they often encounter challenges such as vanishing gradients and high computational overhead due to recursive updates during training. These limitations make RNN-based approaches less favorable compared to their CNN-based and Transformer-based counterparts. Table \ref{LS_CNN} presents CNN-based methods that exploit the strong local feature extraction capabilities of convolutional layers for time series forecasting. While recent models integrating temporal attention, such as ModernTCN \cite{donghao2024moderntcn} and UniRepLKNet \cite{ding2024unireplknet}, have demonstrated notable performance, Transformer-based models have largely outperformed them in terms of scalability and forecasting accuracy, making them the dominant choice in the TSF domain. Table \ref{LS_TSF} illustrates the different transformer-based models proposed for the TSF task. 
\begin{table*}[htbp!]
\centering
\caption{Overview of RNN-based TSF Models}
\label{LS_RNN}
\renewcommand{\arraystretch}{1.1}
\begin{tabular}{|>{\centering\arraybackslash}p{2.3cm}|>{\centering\arraybackslash}p{7cm}|>{\centering\arraybackslash}p{3cm}|>{\centering\arraybackslash}p{3cm}|}
\hline
\textbf{Model} & \textbf{Key Contributions} & \textbf{Advantages} & \textbf{Disadvantages} \\
\hline
LSTNet \cite{lai2018modeling} & Utilizes stacked LSTM layers to model both short- and long-term temporal dependencies effectively in multivariate time series. & Captures both short- and long-term dependencies; effective for multivariate data. & Struggles with irregular time intervals; limited scalability for long sequences. \\
\hline
DA-RNN \cite{qin2017dual} & Enhances forecasting accuracy by incorporating self-attention mechanisms within the RNN to select relevant input features. & Dynamically selects relevant features; improves interpretability. & Attention mechanism adds computational overhead; less effective for noisy data. \\
\hline
Forking-sequences RNN \cite{wen2017multi} & Addresses prediction uncertainty through a forking-sequences training strategy based on an RNN framework. & Handles multi-step prediction uncertainty well; improves robustness. & Requires complex training; sensitive to hyperparameter tuning. \\
\hline
Wavelet-LSTM \cite{wang2018multilevel} & Applies wavelet decomposition to preprocess time series before feeding into an LSTM, capturing both time and frequency domain patterns. & Extracts rich time-frequency features; enhances interpretability. & Additional preprocessing step increases pipeline complexity. \\
\hline
MTNet \cite{chang2018memory} & Combines memory networks with a redesigned attention mechanism to improve long-term forecasting capability. & Strong long-term memory modeling; good for periodic signals. & High memory and computation cost; training may be unstable. \\
\hline
Hybrid-ES-LSTM \cite{smyl2020hybrid} & Integrates exponential smoothing techniques with LSTM architecture to enhance robustness and accuracy. & Combines statistical and deep methods; stable and accurate. & May not generalize well across domains; less flexible for non-stationary data. \\
\hline
TA-Multimodal \cite{fan2019multi} & Leverages a temporal attention mechanism along with multimodal fusion of historical features for improved forecasting performance. & Utilizes multiple data modalities; captures complex interactions. & Requires access to and alignment of multiple modalities; complex architecture. \\
\hline
SegRNN \cite{lin2023segrnn} & Reduces recurrence time through a parallel multi-step prediction strategy, improving the efficiency of RNN-based forecasting. & Improves efficiency via parallelism; suitable for real-time use. & May sacrifice accuracy for speed in some cases; limited to fixed horizons. \\
\hline
WITRAN \cite{jia2023witran} & Builds on SegRNN to model both short- and long-range dependencies, offering improved temporal representation learning. & Balanced modeling of short and long dependencies; better representation. & More complex than SegRNN; may require careful tuning. \\
\hline
\end{tabular}
\end{table*}

\begin{table*}[htbp!]
\centering
\caption{Overview of CNN-based TSF Models}
\label{LS_CNN}
\renewcommand{\arraystretch}{1.1}
\begin{tabular}{|>{\centering\arraybackslash}p{2.3cm}|>{\centering\arraybackslash}p{7cm}|>{\centering\arraybackslash}p{3cm}|>{\centering\arraybackslash}p{3cm}|}
\hline
\textbf{Model} & \textbf{Key Contributions} & \textbf{Advantages} & \textbf{Disadvantages} \\
\hline
DSANet \cite{huang2019dsanet} & Processes each univariate time series independently using dual parallel convolutional layers operating at different temporal scales. & Captures multi-scale temporal patterns effectively, lightweight architecture. & Ignores inter-variable dependencies, limited modeling of multivariate interactions. \\
\hline
SCINet \cite{liu2022scinet} & Introduces a multi-layer binary tree framework to downsample time series into subsequences, modeling both short- and long-term dependencies. & Efficient modeling of hierarchical dependencies; good for long sequences. & Complex architecture; potential loss of resolution during downsampling. \\
\hline
MICN \cite{wang2023micn} & Decomposes time series into trend and seasonal components; integrates representations from multiple trends using CNNs. & Explicit trend-seasonality modeling; interpretable structure. & Requires reliable decomposition; may perform poorly on noisy or non-periodic data. \\
\hline
TimesNet \cite{wu2023timesnet} & Transforms 1D time series into 2D frequency-domain representations using FFT, enabling joint forecasting, classification, and anomaly detection. & Versatile for multiple tasks; exploits frequency-domain correlations. & FFT-based transformation adds computational overhead; sensitivity to frequency noise. \\
\hline
MPPN \cite{wang2023mppn} & Captures multi-resolution and multi-periodic patterns for improved forecasting accuracy using structured convolutional designs. & Handles complex periodic patterns well; strong generalization. & May require extensive tuning of convolutional structure; high training cost. \\
\hline
FDNet \cite{shen2023fdnet} & Combines CNN and linear layers with a focal decomposition strategy to generate representative training subsequences. & Focuses on informative subsequences; good for long sequences. & Risk of discarding valuable temporal information; sensitive to segment quality. \\
\hline
PatchMixer \cite{gong2023patchmixer} & Employs a unified CNN-based architecture to model both short- and long-term temporal dependencies efficiently. & Unified handling of all timescales; efficient and scalable. & May struggle with highly non-stationary data; fixed patch size can be limiting. \\
\hline
ModernTCN \cite{luo2024moderntcn} & Improves CNN-based models by enlarging convolutional kernel size, enhancing the temporal receptive field. & Captures long-range dependencies without attention; low latency. & Large kernels increase parameter count; may overfit on small datasets. \\
\hline
UniRepLKNet \cite{ding2024unireplknet} & Adopts large kernel convolutions to strengthen long-range temporal pattern learning in CNN-based TSF. & Strong performance on long sequences; efficient inference. & Computationally heavy during training; requires careful regularization. \\
\hline
\end{tabular}
\end{table*}

\begin{table*}[htbp!]
\centering
\caption{Overview of Transformer-based TSF Models}
\label{LS_TSF}
\renewcommand{\arraystretch}{1.1}
\begin{tabular}{|>{\centering\arraybackslash}p{2.3cm}|>{\centering\arraybackslash}p{7cm}|>{\centering\arraybackslash}p{3cm}|>{\centering\arraybackslash}p{3cm}|}
\hline
\textbf{Model} & \textbf{Key Contributions} & \textbf{Advantages} & \textbf{Disadvantages} \\
\hline
Informer \cite{zhou2021informer} & Efficiently models long-range temporal dependencies using a sparse self-attention mechanism tailored for time series forecasting. & Reduces attention complexity; scalable to long sequences. & May underperform on short sequences or irregular time steps. \\
\hline
Autoformer \cite{wu2021autoformer} & Introduces an autocorrelation-based decomposition to capture segment-level periodicities in temporal sequences. & Models periodic patterns effectively; avoids redundancy. & Assumes inherent periodicity; less effective for non-repeating patterns. \\
\hline
PatchTST \cite{hochreiter1997long} & Splits time series into fixed-length patches, enabling parallel learning of local and global patterns. & Enables efficient training; captures multiple scales. & Fixed patch size may miss important transitions; requires careful tuning. \\
\hline
Crossformer \cite{zhang2023crossformer} & Employs a dual-path attention mechanism to integrate both fine-grained and coarse-grained temporal dependencies. & Balanced short- and long-term modeling; flexible design. & Computationally intensive due to dual-path structure. \\
\hline
BasisFormer \cite{ni2023basisformer} & Enhances forecasting through adaptive self-supervised learning and bidirectional cross-attention across time steps. & Leverages self-supervised learning; bidirectional context use. & Increased architectural complexity; slower inference. \\
\hline
iTransformer \cite{liu2023itransformer} & Learns variate-centric representations by inverting input dimensions and applying Transformer blocks accordingly. & Strong performance on multivariate data; emphasizes variable relationships. & Inversion strategy may hinder scalability on very high-dimensional data. \\
\hline
CATN \cite{he2022catn} & Utilizes a tree-hierarchical structure to capture multi-scale dependencies across global and local contexts. & Good hierarchical learning; suitable for structured signals. & Tree-based structure may complicate training and optimization. \\
\hline
CACRN \cite{bai2023cluster} & Incorporates variable clustering and separate processing of dependent vs. independent variables for improved TSF performance. & Effective feature separation; improved multivariate forecasting. & Requires reliable clustering; susceptible to cluster quality. \\
\hline
MrCAN \cite{zhang2023mrcan} & Applies batch-level attention to exploit cross-sample correlations, enhancing forecasting in sparse training scenarios. & Robust in low-data regimes; utilizes inter-sample signals. & Assumes inter-sample similarity; vulnerable to noise across samples. \\
\hline
MTCAN \cite{wan2022multivariate} & Combines dilated convolutions with attention layers to strengthen both short- and long-range temporal modeling. & Efficient temporal modeling; improves receptive field. & Fusion of modules increases design complexity. \\
\hline
VD-Triangle \cite{he2023multi} & Integrates variable distillation attention with triangle structure learning to improve multivariate forecasting accuracy. & Enhances variable interaction learning; promotes interpretability. & High model complexity; slower training time. \\
\hline
\end{tabular}
\end{table*}
\section{Methodology}
This section begins by outlining the fundamental concepts underlying the Transformer and Daubechies Wavelets coefficients, and subsequently details the proposed DB2-TransF architecture tailored for the time series forecasting (TSF) task.

\subsection{\textbf{Preliminaries}}
\subsubsection{\textbf{Transformer}}

The working of the traditional Transformer proposed by Vaswani et al.~\cite{vaswani2017attention} can be summarized as follows:
\begin{itemize}
\item Initially, the tokenization step is performed to generate the tokens from the given input.  Consider an input sequence of $T$ tokens. Where each token is embedded into a $d$-dimensional vector and forms an input matrix $\bm{Z} \in \mathbb{R}^{T \times d}$.

\item The input $Z$ obtained above is  linearly projected into three distinct spaces to obtain:
    \begin{align}
        \bm{A} &= \bm{Z} \cdot \bm{W^A}, \\
        \bm{B} &= \bm{Z} \cdot \bm{W^B}, \\
        \bm{C} &= \bm{Z} \cdot \bm{W^C},
    \end{align}
    Where: $\bm{A} \in \mathbb{R}^{T \times d_k}$ is the query matrix, $\bm{B} \in \mathbb{R}^{T \times d_k}$ is the key matrix, $\bm{C} \in \mathbb{R}^{T \times d_v}$ is the value matrix, $\bm{W^A}, \bm{W^B} \in \mathbb{R}^{d \times d_k}$ and $\bm{W^C} \in \mathbb{R}^{d \times d_v}$ are trainable weight matrices.

    \item 
    The attention weights is computed by taking the dot product of queries and keys, scaled by $\sqrt{d_k}$ (where $d_k$ is the key/query dimension) to stabilize gradients, and normalized using the softmax function ($\sigma$):
    \begin{equation}
        \text{Attention}(\bm{A}, \bm{B}, \bm{C}) = \sigma \left( \frac{\bm{A} \cdot \bm{B}^\top}{\sqrt{d_k}} \right) \cdot \bm{C}
    \end{equation}
    This produces the attention matrix $ \bm{Z}_{\text{att}}$ of shape $\mathbb{R}^{T \times d_v}$ that represents how much each token attends to the others and also captures the contextualized representation of each token based on the full sequence.

    \item The obtained attention output $ \bm{Z}_{\text{att}}$ is passed through a linear transformation for integration with subsequent network layers:
    \begin{equation}
        \bm{Y} = \bm{Z}_{\text{att}} \cdot \bm{W}^P
    \end{equation}
    where $\bm{W}^P \in \mathbb{R}^{d_v \times d}$ is a learnable output projection matrix and $\bm{Y} \in \mathbb{R}^{T \times d}$ is the final output of the attention block.

    \item 
    The steps outlined above, carried out by the transformer's attention block, enable the model to dynamically assess the relevance of each token with respect to others in the sequence. This facilitates the modeling of long-range dependencies. Moreover, the flexibility to handle inputs of varying lengths and to learn contextual relationships through attention mechanisms is central to the effectiveness of transformer-based architectures across a wide range of sequential and multi-modal tasks.
\end{itemize}

\subsection{\textbf{Classical Daubechies (DB2) Wavelet Transform }}
The Daubechies (DB2) wavelet transform is a popular wavelet transform method employed in the domain of signal processing and image compression, particularly for denoising applications. The DB2 wavelet efficiently captures the various frequency components in a signal and images, which makes it a preferred choice for feature extraction. Additionally, the DB2 wavelet transform is computationally efficient compared to other wavelet transforms and employs low and high-pass filters to capture the small and noise-related features from a signal or an image. 
The various steps involved while applying the  DB2 wavelet transform are as follows: 

\begin{enumerate}
    \item For a one-dimensional signal $x$ of even length $n$. The DB2 wavelet transform computes two sets of coefficients by convolving the low-pass filter $h_{f}$ and the high-pass filter $g_{f}$ across the $x$. The values of four coefficients of $h_{f}$  and $g_{f}$ are as follows:  

    \begin{equation}\label{Low_Filter_Coff}
        h_{f}[0]=\frac{1+\sqrt{3}}{4\sqrt{2}}, h_{f}[1]=\frac{3+\sqrt{3}}{4\sqrt{2}}, h_{f}[2]=\frac{3-\sqrt{3}}{4\sqrt{2}}, h_{f}[3]=\frac{1-\sqrt{3}}{4\sqrt{2}}
    \end{equation}

     \begin{equation} \label{High_Filter_Coff}
       g_{f}[0]=\frac{-1+\sqrt{3}}{4\sqrt{2}}, g_{f}[1]=\frac{3-\sqrt{3}}{4\sqrt{2}},  g_{f}[2]=\frac{-3-\sqrt{3}}{4\sqrt{2}}, g_{f}[3]=\frac{1+\sqrt{3}}{4\sqrt{2}}
    \end{equation}
    \item The $h_{f}$ filter is employed to extract the approximation coefficients $(A)$ from the time signal, capturing the slow variations or smooth trends across the time series. The computation of the approximation coefficients is given by: 
    \begin{equation}
        A[n]=\sum_{k=0}^{3}h_{f}[k] \odot x[2n+k]
    \end{equation}
    Here, $k$ denotes the $k^{th}$ low filter coefficient, and $n$ represents the total number of approximation coefficients computed using the filter $h_{f}$. Here, $\odot$ denotes element-wise multiplication.

    \item The $g_{f}$ filter is employed to extract the detailed coefficients $(D)$ from the time signal, capturing the fast variations and noise across the time series. The computation of the detailed coefficients is given by:
    \begin{equation}
           D[n]=\sum_{k=0}^{3}g_{f}[k] \odot x[2n+k]
    \end{equation}
     Here, $k$ denotes the $k^{th}$ high filter coefficient, and $n$ represents the total number of detailed coefficients computed using the filter $g_{f}$.
\end{enumerate}
\subsection{\textbf{Proposed Method}}
This section outlines the construction of learnable DB2 coefficients, the design of learnable multiscale DB2 filters, and the overall architecture of DB2-TransF along with its various layers. 
\subsubsection{\textbf{Learnable Daubechies (DB2) Wavelet Module}}

The coefficients of the low-pass and high-pass filters, denoted by $\alpha^{k} \in \mathbb{R}^{d}$ and $\beta^{k} \in \mathbb{R}^{d}$ respectively, are made learnable by initializing them with values close to their corresponding predefined coefficients, as shown in Eq.(\ref{Low_Filter_Coff}) and Eq.(\ref{High_Filter_Coff}). For instance, $ \alpha^{0}\approx\frac{1+\sqrt{3}}{4\sqrt{2}}, \alpha^{1}\approx\frac{3+\sqrt{3}}{4\sqrt{2}}, \alpha^{2}\approx\frac{3-\sqrt{3}}{4\sqrt{2}}, \alpha^{3}\approx\frac{1-\sqrt{3}}{4\sqrt{2}},   \beta^{0}\approx\frac{-1+\sqrt{3}}{4\sqrt{2}}, \beta^{1}\approx\frac{3-\sqrt{3}}{4\sqrt{2}},  \beta^{2}\approx\frac{-3-\sqrt{3}}{4\sqrt{2}}, \beta^{3}\approx\frac{1+\sqrt{3}}{4\sqrt{2}}$.
These $d$-dimensional learnable coefficients are optimized via backpropagation. The approximate and detailed coefficients, computed using the learnable $\alpha^{k}$ and $\beta^{k}$, are obtained through Eq.(\ref{Learnable_Approximate_Value}) and Eq.(\ref{Learnable_Detailed_Value}), respectively. 
\begin{equation}\label{Learnable_Approximate_Value}
    A[n]=\sum_{k=0}^{3} \alpha^{k} \odot x[2n+k]
\end{equation}
\begin{equation}\label{Learnable_Detailed_Value}
    D[n]=\sum_{k=0}^{3}\beta^{k} \odot x[2n+k]
\end{equation}
\subsubsection{\textbf{Multi-Scale Learnable Daubechies (DB2) Wavelet Module}}

At level 0, the computation of approximate and detailed coefficients is performed using Eq.(\ref{Learnable_Approximate_Value}) and Eq.(\ref{Learnable_Detailed_Value}), respectively. The computation of the same set of coefficients at the level $l$ is computed using Eq.(\ref{Learnable_Approximate_Value1}) and Eq.(\ref{Learnable_Detailed_Value1}), respectively.

\begin{equation}\label{Learnable_Approximate_Value1}
    A_{l}[n]=\sum_{k=0}^{3} \alpha^{k}_{l} \odot x_{l}[2n+k]
\end{equation}
\begin{equation}\label{Learnable_Detailed_Value1}
    D_{l}[n]=\sum_{k=0}^{3}\beta^{k}_{l} \odot x_{l}[2n+k]
\end{equation}
For the level $l+1$, the input $x$ is set to the approximation coefficient of the previous level $l$. 
\begin{equation}
    x_{l+1}=A_{l}
\end{equation}
This process generates a sequence of detail coefficients $[{D_{1}, \ldots, D_{L}}]$ at multiple scales or resolutions, along with the final approximation coefficient sequence $A_{L}$. Together, these constitute the multiscale decomposition of the original input $x$. The final representation of $x$ in terms of its approximate and detailed coefficients is given by:

\begin{equation}
    x=\left\{A_{L},D_{1}, \ldots, D_{L}\right\}
\end{equation}

\subsubsection{\textbf{Proposed DB2-TransF Architecture}}
\begin{figure}[htbp!]
    \centering
    \includegraphics[width=13cm]{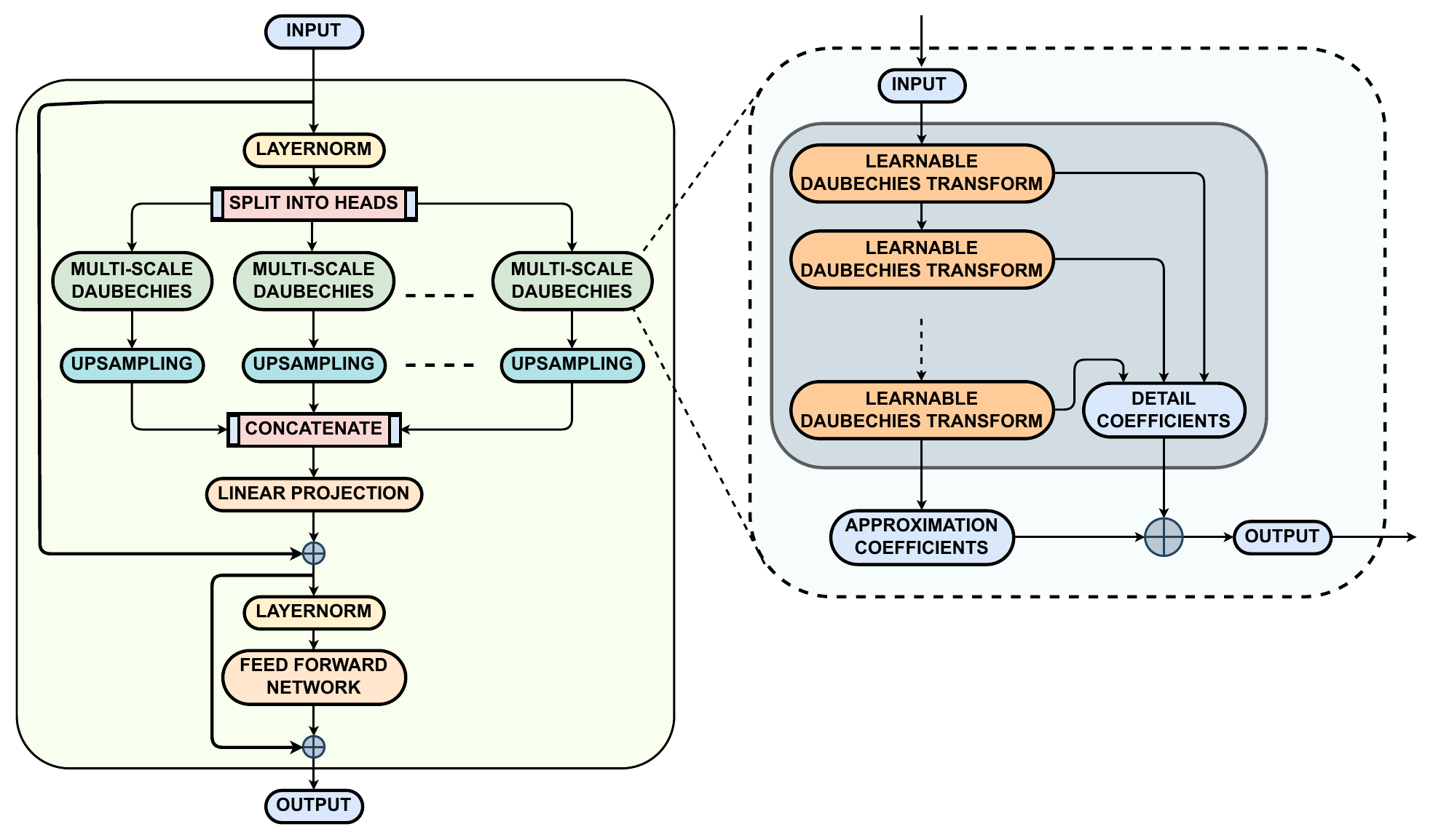}
    \caption{Architecture of DB2-Transformer With Multi-Scale Learnable  Daubechies (DB2) Wavelet Block}
    \label{DB2_Trans_Architecture}
\end{figure}

Figure \ref{DB2_Trans_Architecture} illustrates the overall architecture of the proposed DB2-TransF crafted for the time series classification. The newly designed multiscale learnable Daubechies wavelet block (MLDB) makes the DB2-TransF effectively extract the approximate and detailed coefficients from the input time series while ensuring efficient learning of correlations between multiple variables. Algorithm \ref{DB2_algo} illustrates the overall working of the proposed DB2-TransF Model.

\begin{algorithm}[H]
\caption{DB2-Transformer for Time Series Forecasting}
\label{DB2_algo}
\begin{algorithmic}[1]
\Require Input Time Series $x \in \mathbb{R}^{B \times T \times D}$ \Comment{$B$: Batch Size, $T$: Input Sequence Length, $D$: Feature Dimension After Linear Projection}
\Require Parameters: Levels $L$, Heads $H$, Depth $N$, Forecast Horizon $T_{\text{pred}}$
\Ensure Forecasted Output $\hat{y} \in \mathbb{R}^{B \times T_{\text{pred}} \times D}$


\For{$i = 1$ to $N$} \textcolor{blue}{\Comment{$N$: Number of MLDB Blocks}}
    \State $x \gets \text{LayerNorm}(x)$ \textcolor{blue}{\Comment{Apply Layer Normalization for Stabilize Training}}

    \Statex \textbf{Step 2: Multi-Head Multi-Scale Daubechies Transform}
    \State Split $x$ into $H$ Heads: $\{x^{(1)}, \dots, x^{(H)}\}$ \textcolor{blue}{\Comment{Each $x^{(h)} \in \mathbb{R}^{B \times T \times D/H}$}}

    \For{$h = 1$ to $H$} \textcolor{blue}{\Comment{$H$: Number of Heads}}
        \State $x^{(h)}_0 \gets x^{(h)}$ \textcolor{blue}{\Comment{Input for DB2 Wavelet Head $h$ at Level 0}}
        \For{$l = 1$ to $L$} \textcolor{blue}{\Comment{$L$: number of wavelet decomposition levels}}
            \State Padding is Performed Over $x^{(h)}_{l-1}$, If Needed to Ensure Windowed Wavelet Convolution Works
            \State Unfold $x^{(h)}_{l-1}$ into Windows of Size $W_{s}$ With Stride $r$ \textcolor{blue}{\Comment{Typically, $W_{s}$ and $r$ Are Chosen As Multiples of 2, Since The number of Coefficients in Both The Low-Pass and High-Pass Learnable Filters Is Even.}}
            \State Apply Learnable Daubechies Filters $\alpha^{k(h)}_{l}, \beta^{k(h)}_{l}$:
            \[
            A_l^{(h)} = \sum_{k=0}^{3} \alpha^{k(h)}_{l} \odot x^{(h_{t+k})}_{l}, \quad
            D_l^{(h)} = \sum_{k=0}^{3} \beta^{k(h)}_{l} \odot x^{(h_{t+k})}_{l}
            \]
           \textcolor{blue}{ \Comment{$A_l$: Low-Pass (Approximation); $D_l$: High-Pass (Detailed) Coefficients}}
            \State $x^{(h)}_{l+1} \gets A_l^{(h)}$ \textcolor{blue}{\Comment{Use Approximation For Next Level}}
        \EndFor
        \State $O^{(h)} \gets \text{Concatenate}(A_{L}, D_1, \dots, D_L)$ \textcolor{blue}{\Comment{Final Output of Head $h$}}
    \EndFor

    \State Concatenate All Heads: $O \gets \text{Concatenate}(O^{(1)}, \dots, O^{(H)})$ Along Feature Axis
    \State Project Back to Input Dimension: $O \gets \text{Linear}(O)$ \textcolor{blue}{\Comment{$O \in \mathbb{R}^{B \times T' \times D}$}}
    \State Align Temporal Length: $x_{\text{aligned}} \gets \text{Trim}(x)$ to Match Dimension of $O$
    \State Add Residual Connection: $x \gets x_{\text{aligned}} + O$ \textcolor{blue}{\Comment{Residual Connection}}

    \Statex \textbf{Step 3: Feedforward Network}
    \State $x \gets x + \text{FFN}(\text{LayerNorm}(x))$ \textcolor{blue}{\Comment{FFN with Two MLP Layers and GELU Activation}}
\EndFor

\Statex \textbf{Step 4: Forecasting Head}
\State $\hat{y} \gets \text{ForecastHead}(x)$ \textcolor{blue}{\Comment{Maps Encoded Sequence to Future Outputs: $\hat{y} \in \mathbb{R}^{B \times T_{\text{pred}} \times D}$}}

\end{algorithmic}
\end{algorithm}
\section{Experiments and Results}

This section provides detailed information on the experimental setup, datasets, results, ablation studies, and related discussions.
\subsection{\textbf{Experimental Setup}}
Table \ref{tab:exp_setup} presents the details of the experimental setup employed in all the experiments conducted in this study.
 \begin{table}[htbp]
\centering
\caption{Experimental Setup and Hyperparameters}
\label{tab:exp_setup}
\renewcommand{\arraystretch}{1.4}
\scalebox{0.70}{
\begin{tabular}{|c|c|}
\hline
\textbf{Component}               & \textbf{Details}      \\ \hline
\textbf{Optimizer}               & AdamW                 \\ \hline
\textbf{Learning Rate Scheduler} & Expnential Decay      \\ \hline
\textbf{Early Stopping Patience} & 5 epochs              \\ \hline
\textbf{Programming Language}    & Python 3.10.16        \\ \hline
\textbf{CUDA Version}            & 12.1                  \\ \hline
\textbf{Operating System}        & Ubuntu 24.04.1 LTS    \\ \hline
\textbf{Hardware}                & RTX 4060 GPU (single) \\ \hline
\end{tabular}}
\end{table}
\subsection{\textbf{Compared Models}}
Table \ref{tab:model_summary} lists the recent models compared against the proposed model for performance evaluation.
\begin{table}[htbp!]
\centering
\caption{Summary of Compared Time Series Forecasting Models}
\scalebox{0.75}{
\begin{tabular}{|c|c|p{9cm}|}
\hline
\textbf{Model} & \textbf{Category} & \textbf{Key Characteristics} \\ \hline
\textbf{iTransformer} \cite{liu2023itransformer} & Transformer-based & Processes each variate independently prior to multivariate fusion and is regarded as the current state-of-the-art (SOTA) in time series forecasting.\\ \hline
\textbf{PatchTST} \cite{nie2022patchtst} & Transformer-based & Divides the time series into patches and applies channel-independent shared embeddings and weights for feature extraction. \\ \hline
\textbf{Crossformer} \cite{zhang2023crossformer} & Transformer-based & Utilizes cross-attention mechanisms to effectively capture long-range dependencies across temporal sequences. \\ \hline
\textbf{FEDformer} \cite{zhou2022fedformer} & Transformer-based & Improves Transformer performance by leveraging frequency-domain sparsity, typically through Fourier transforms. \\ \hline
\textbf{Autoformer} \cite{wu2021autoformer} & Transformer-based & Employs a decomposition-based architecture combined with an auto-correlation mechanism for effective time series modeling. \\ \hline
\textbf{RLinear} \cite{zhang2023rlinear} & Linear-based & A state-of-the-art linear model that incorporates reversible normalization and assumes channel independence.\\ \hline
\textbf{TiDE} \cite{cirstea2023tide} & Linear-based & An encoder–decoder architecture built entirely using multi-layer perceptrons (MLPs).\\ \hline
\textbf{DLinear} \cite{zeng2022dlinear} & Linear-based & Among the earliest linear models for time series forecasting, utilizing a single-layer architecture combined with series decomposition.\\ \hline
\textbf{TimesNet} \cite{wu2023timesnet} & Temporal Conv-based & Employs 2D convolutional kernels (TimesBlock) to model both intra-period and inter-period variations in time series data.\\ \hline
\end{tabular}}
\label{tab:model_summary}
\end{table}
\subsection{\textbf{Dataset Description}}
Table \ref{dataset} provides a description of the various datasets used to evaluate the performance of the proposed model for time series forecasting.
\begin{table*}[htbp!]
\centering
\caption{Summary and Statistics of the Thirteen Public TSF Datasets Used for Experiments}
\label{dataset}
\scalebox{0.75}{
\begin{tabular}{|p{0.8cm}|p{3cm}|p{5.5cm}|c|c|c|}
\hline
\textbf{\rotatebox{90}{Category}} & \textbf{Dataset(s)} & \textbf{Description} & \textbf{Variates} & \textbf{Timesteps} & \textbf{Granularity} \\
\hline
\multirow{2}{*}{\rotatebox{90}{Traffic}} 
& Traffic \cite{wu2021autoformer} & Hourly road occupancy data from 862 sensors in the San Francisco Bay Area (2015–2016); highly multivariate and periodic. & 862 & 17,544 & 1 h \\ 
& PEMS03/04/07/08  \cite{chen2001freeway}& Spatiotemporal traffic flow from California highway systems; widely used, highly multivariate, and periodic. & 358 / 307 / 883 / 170 & 26,209 / 16,992 / 28,224 / 17,856 & 5 min \\
\hline
\multirow{2}{*}{\rotatebox{90}{ETT \cite{zhou2021informer}}} 
& ETTm1, ETTm2, ETTh1, ETTh2 & Transformer load and oil temperature records (2016–2018); low-variate, weakly regular. & 7 & 17,420 / 69,680 & 15 min / 1 h \\
\hline
\multirow{4}{*}{\rotatebox{90}{Other}} 
& Electricity \cite{wu2021autoformer} & Hourly electricity usage of 321 customers (2012–2014); strongly periodic and multivariate. & 321 & 26,304 & 1 h \\
& Exchange \cite{wu2021autoformer}& Daily exchange rates of 8 countries (1990–2016); low-variate and mostly aperiodic. & 8 & 7,588 & 1 day \\
& Weather \cite{wu2021autoformer} & 21 meteorological features every 10 min from a weather station in 2020; sparse and largely aperiodic. & 21 & 52,696 & 10 min \\
& Solar-Energy \cite{lai2018modeling} & Solar power generation from 137 PV plants in Alabama (2006); multivariate and highly periodic. & 137 & 52,560 & 10 min \\
\hline
\end{tabular}}
\end{table*}

\section{Results}
This section presents the results, ablation study, and discussion. A comparative evaluation of the models is carried out using two primary performance metrics: the average Mean Squared Error (MSE) and Mean Absolute Error (MAE) computed across different sequence lengths as used in \cite{liu2023itransformer}.

\subsection{\textbf{Performance Comparison}}
Tables \ref{Dataset-1}, \ref{Dataset-2}, and \ref{Dataset-3} present a comparative analysis of the proposed DB2-TransF model against state-of-the-art methods for time series forecasting (TSF) across the datasets listed in Table \ref{dataset}. 

\begin{sidewaystable}
\caption{Performance Comparison of DB2-TransF and Other State-of-the-Art Methods for Electricity, Weather, Solar and Exchange Datasets}
\label{Dataset-1}
\centering 
\scalebox{0.60}{
\begin{tabular}{|cc|cc|cc|cc|cc|cc|cc|cc|cc|cc|cc|cc|cc|cc|}
\hline
\multicolumn{2}{|c|}{\textbf{Models}}                                                                 & \multicolumn{2}{c|}{\textbf{DB2-TransF}}             & \multicolumn{2}{c|}{\textbf{iTransformer}}           & \multicolumn{2}{c|}{\textbf{RLinear}}                & \multicolumn{2}{c|}{\textbf{PatchTST}}               & \multicolumn{2}{c|}{\textbf{Crossformer}}            & \multicolumn{2}{c|}{\textbf{TiDE}}                   & \multicolumn{2}{c|}{\textbf{TimesNet}}               & \multicolumn{2}{c|}{\textbf{DLinear}}                & \multicolumn{2}{c|}{\textbf{FEDformer}}              & \multicolumn{2}{c|}{\textbf{Autoformer}}             \\ \hline
\multicolumn{2}{|c|}{\textbf{Metric}}                                                                 & \multicolumn{1}{c|}{\textbf{MSE}}   & \textbf{MAE}   & \multicolumn{1}{c|}{\textbf{MSE}}   & \textbf{MAE}   & \multicolumn{1}{c|}{\textbf{MSE}}   & \textbf{MAE}   & \multicolumn{1}{c|}{\textbf{MSE}}   & \textbf{MAE}   & \multicolumn{1}{c|}{\textbf{MSE}}   & \textbf{MAE}   & \multicolumn{1}{c|}{\textbf{MSE}}   & \textbf{MAE}   & \multicolumn{1}{c|}{\textbf{MSE}}   & \textbf{MAE}   & \multicolumn{1}{c|}{\textbf{MSE}}   & \textbf{MAE}   & \multicolumn{1}{c|}{\textbf{MSE}}   & \textbf{MAE}   & \multicolumn{1}{c|}{\textbf{MSE}}   & \textbf{MAE}   \\ \hline
\multicolumn{1}{|c|}{\multirow{5}{*}{\rotatebox[origin=c]{90}{\textbf{Electricity}}}}  & \textbf{96}  & \multicolumn{1}{c|}{0.148}          & 0.243          & \multicolumn{1}{c|}{0.148}          & 0.240          & \multicolumn{1}{c|}{0.201}          & 0.281          & \multicolumn{1}{c|}{0.181}          & 0.270          & \multicolumn{1}{c|}{0.219}          & 0.314          & \multicolumn{1}{c|}{0.237}          & 0.329          & \multicolumn{1}{c|}{0.168}          & 0.272          & \multicolumn{1}{c|}{0.197}          & 0.282          & \multicolumn{1}{c|}{0.193}          & 0.308          & \multicolumn{1}{c|}{0.201}          & 0.317          \\ \cline{2-22} 
\multicolumn{1}{|c|}{}                                                                 & \textbf{192} & \multicolumn{1}{c|}{0.163}          & 0.256          & \multicolumn{1}{c|}{0.162}          & 0.253          & \multicolumn{1}{c|}{0.201}          & 0.283          & \multicolumn{1}{c|}{0.188}          & 0.274          & \multicolumn{1}{c|}{0.231}          & 0.322          & \multicolumn{1}{c|}{0.236}          & 0.330          & \multicolumn{1}{c|}{0.184}          & 0.289          & \multicolumn{1}{c|}{0.196}          & 0.285          & \multicolumn{1}{c|}{0.201}          & 0.315          & \multicolumn{1}{c|}{0.222}          & 0.334          \\ \cline{2-22} 
\multicolumn{1}{|c|}{}                                                                 & \textbf{336} & \multicolumn{1}{c|}{0.182}          & 0.275          & \multicolumn{1}{c|}{0.178}          & 0.269          & \multicolumn{1}{c|}{0.215}          & 0.298          & \multicolumn{1}{c|}{0.204}          & 0.293          & \multicolumn{1}{c|}{0.246}          & 0.337          & \multicolumn{1}{c|}{0.249}          & 0.344          & \multicolumn{1}{c|}{0.198}          & 0.300          & \multicolumn{1}{c|}{0.209}          & 0.301          & \multicolumn{1}{c|}{0.214}          & 0.329          & \multicolumn{1}{c|}{0.231}          & 0.338          \\ \cline{2-22} 
\multicolumn{1}{|c|}{}                                                                 & \textbf{720} & \multicolumn{1}{c|}{0.212}          & 0.307          & \multicolumn{1}{c|}{0.225}          & 0.317          & \multicolumn{1}{c|}{0.257}          & 0.331          & \multicolumn{1}{c|}{0.246}          & 0.324          & \multicolumn{1}{c|}{0.280}          & 0.363          & \multicolumn{1}{c|}{0.284}          & 0.373          & \multicolumn{1}{c|}{0.220}          & 0.320          & \multicolumn{1}{c|}{0.245}          & 0.333          & \multicolumn{1}{c|}{0.246}          & 0.355          & \multicolumn{1}{c|}{0.254}          & 0.361          \\ \cline{2-22} 
\multicolumn{1}{|c|}{}                                                                 & \textbf{Avg} & \multicolumn{1}{c|}{\textbf{0.176}} & \textbf{0.270} & \multicolumn{1}{c|}{\textbf{0.178}} & \textbf{0.270} & \multicolumn{1}{c|}{\textbf{0.219}} & \textbf{0.298} & \multicolumn{1}{c|}{\textbf{0.205}} & \textbf{0.290} & \multicolumn{1}{c|}{\textbf{0.244}} & \textbf{0.334} & \multicolumn{1}{c|}{\textbf{0.251}} & \textbf{0.344} & \multicolumn{1}{c|}{\textbf{0.192}} & \textbf{0.295} & \multicolumn{1}{c|}{\textbf{0.212}} & \textbf{0.300} & \multicolumn{1}{c|}{\textbf{0.214}} & \textbf{0.327} & \multicolumn{1}{c|}{\textbf{0.227}} & \textbf{0.338} \\ \hline
\multicolumn{1}{|c|}{\multirow{5}{*}{\rotatebox[origin=c]{90}{\textbf{Exchange}}}}     & \textbf{96}  & \multicolumn{1}{c|}{0.083}          & 0.203          & \multicolumn{1}{c|}{0.086}          & 0.206          & \multicolumn{1}{c|}{0.093}          & 0.217          & \multicolumn{1}{c|}{0.088}          & 0.205          & \multicolumn{1}{c|}{0.256}          & 0.367          & \multicolumn{1}{c|}{0.094}          & 0.218          & \multicolumn{1}{c|}{0.107}          & 0.234          & \multicolumn{1}{c|}{0.088}          & 0.218          & \multicolumn{1}{c|}{0.148}          & 0.278          & \multicolumn{1}{c|}{0.197}          & 0.323          \\ \cline{2-22} 
\multicolumn{1}{|c|}{}                                                                 & \textbf{192} & \multicolumn{1}{c|}{0.169}          & 0.294          & \multicolumn{1}{c|}{0.177}          & 0.299          & \multicolumn{1}{c|}{0.184}          & 0.307          & \multicolumn{1}{c|}{0.176}          & 0.299          & \multicolumn{1}{c|}{0.470}          & 0.509          & \multicolumn{1}{c|}{0.184}          & 0.307          & \multicolumn{1}{c|}{0.226}          & 0.344          & \multicolumn{1}{c|}{0.176}          & 0.315          & \multicolumn{1}{c|}{0.271}          & 0.315          & \multicolumn{1}{c|}{0.300}          & 0.369          \\ \cline{2-22} 
\multicolumn{1}{|c|}{}                                                                 & \textbf{336} & \multicolumn{1}{c|}{0.311}          & 0.404          & \multicolumn{1}{c|}{0.331}          & 0.417          & \multicolumn{1}{c|}{0.351}          & 0.432          & \multicolumn{1}{c|}{0.301}          & 0.397          & \multicolumn{1}{c|}{1.268}          & 0.883          & \multicolumn{1}{c|}{0.349}          & 0.431          & \multicolumn{1}{c|}{0.367}          & 0.448          & \multicolumn{1}{c|}{0.313}          & 0.427          & \multicolumn{1}{c|}{0.460}          & 0.427          & \multicolumn{1}{c|}{0.509}          & 0.524          \\ \cline{2-22} 
\multicolumn{1}{|c|}{}                                                                 & \textbf{720} & \multicolumn{1}{c|}{0.777}          & 0.662          & \multicolumn{1}{c|}{0.847}          & 0.691          & \multicolumn{1}{c|}{0.886}          & 0.714          & \multicolumn{1}{c|}{0.901}          & 0.714          & \multicolumn{1}{c|}{1.767}          & 1.068          & \multicolumn{1}{c|}{0.852}          & 0.698          & \multicolumn{1}{c|}{0.964}          & 0.746          & \multicolumn{1}{c|}{0.839}          & 0.695          & \multicolumn{1}{c|}{1.195}          & 0.695          & \multicolumn{1}{c|}{1.447}          & 0.941          \\ \cline{2-22} 
\multicolumn{1}{|c|}{}                                                                 & \textbf{Avg} & \multicolumn{1}{c|}{\textbf{0.335}} & \textbf{0.391} & \multicolumn{1}{c|}{\textbf{0.360}} & \textbf{0.403} & \multicolumn{1}{c|}{\textbf{0.378}} & \textbf{0.417} & \multicolumn{1}{c|}{\textbf{0.367}} & \textbf{0.404} & \multicolumn{1}{c|}{\textbf{0.940}} & \textbf{0.707} & \multicolumn{1}{c|}{\textbf{0.370}} & \textbf{0.413} & \multicolumn{1}{c|}{\textbf{0.416}} & \textbf{0.443} & \multicolumn{1}{c|}{\textbf{0.354}} & \textbf{0.414} & \multicolumn{1}{c|}{\textbf{0.519}} & \textbf{0.429} & \multicolumn{1}{c|}{\textbf{0.613}} & \textbf{0.539} \\ \hline
\multicolumn{1}{|c|}{\multirow{5}{*}{\rotatebox[origin=c]{90}{\textbf{Weather}}}}      & \textbf{96}  & \multicolumn{1}{c|}{0.165}          & 0.210          & \multicolumn{1}{c|}{0.174}          & 0.214          & \multicolumn{1}{c|}{0.192}          & 0.232          & \multicolumn{1}{c|}{0.177}          & 0.218          & \multicolumn{1}{c|}{0.158}          & 0.230          & \multicolumn{1}{c|}{0.202}          & 0.261          & \multicolumn{1}{c|}{0.172}          & 0.220          & \multicolumn{1}{c|}{0.196}          & 0.255          & \multicolumn{1}{c|}{0.217}          & 0.296          & \multicolumn{1}{c|}{0.266}          & 0.336          \\ \cline{2-22} 
\multicolumn{1}{|c|}{}                                                                 & \textbf{192} & \multicolumn{1}{c|}{0.218}          & 0.257          & \multicolumn{1}{c|}{0.221}          & 0.254          & \multicolumn{1}{c|}{0.240}          & 0.271          & \multicolumn{1}{c|}{0.225}          & 0.259          & \multicolumn{1}{c|}{0.206}          & 0.277          & \multicolumn{1}{c|}{0.242}          & 0.298          & \multicolumn{1}{c|}{0.219}          & 0.261          & \multicolumn{1}{c|}{0.237}          & 0.296          & \multicolumn{1}{c|}{0.276}          & 0.336          & \multicolumn{1}{c|}{0.307}          & 0.367          \\ \cline{2-22} 
\multicolumn{1}{|c|}{}                                                                 & \textbf{336} & \multicolumn{1}{c|}{0.272}          & 0.295          & \multicolumn{1}{c|}{0.278}          & 0.296          & \multicolumn{1}{c|}{0.292}          & 0.307          & \multicolumn{1}{c|}{0.278}          & 0.297          & \multicolumn{1}{c|}{0.272}          & 0.335          & \multicolumn{1}{c|}{0.287}          & 0.335          & \multicolumn{1}{c|}{0.280}          & 0.306          & \multicolumn{1}{c|}{0.283}          & 0.335          & \multicolumn{1}{c|}{0.339}          & 0.380          & \multicolumn{1}{c|}{0.359}          & 0.395          \\ \cline{2-22} 
\multicolumn{1}{|c|}{}                                                                 & \textbf{720} & \multicolumn{1}{c|}{0.352}          & 0.348          & \multicolumn{1}{c|}{0.358}          & 0.347          & \multicolumn{1}{c|}{0.364}          & 0.353          & \multicolumn{1}{c|}{0.354}          & 0.348          & \multicolumn{1}{c|}{0.398}          & 0.418          & \multicolumn{1}{c|}{0.351}          & 0.386          & \multicolumn{1}{c|}{0.365}          & 0.359          & \multicolumn{1}{c|}{0.345}          & 0.381          & \multicolumn{1}{c|}{0.403}          & 0.428          & \multicolumn{1}{c|}{0.419}          & 0.428          \\ \cline{2-22} 
\multicolumn{1}{|c|}{}                                                                 & \textbf{Avg} & \multicolumn{1}{c|}{\textbf{0.252}} & \textbf{0.278} & \multicolumn{1}{c|}{\textbf{0.258}} & \textbf{0.278} & \multicolumn{1}{c|}{\textbf{0.272}} & \textbf{0.291} & \multicolumn{1}{c|}{\textbf{0.259}} & \textbf{0.281} & \multicolumn{1}{c|}{\textbf{0.259}} & \textbf{0.315} & \multicolumn{1}{c|}{\textbf{0.271}} & \textbf{0.320} & \multicolumn{1}{c|}{\textbf{0.259}} & \textbf{0.287} & \multicolumn{1}{c|}{\textbf{0.265}} & \textbf{0.317} & \multicolumn{1}{c|}{\textbf{0.309}} & \textbf{0.360} & \multicolumn{1}{c|}{\textbf{0.338}} & \textbf{0.382} \\ \hline
\multicolumn{1}{|c|}{\multirow{5}{*}{\rotatebox[origin=c]{90}{\textbf{Solar-Energy}}}} & \textbf{96}  & \multicolumn{1}{c|}{0.224}          & 0.265          & \multicolumn{1}{c|}{0.203}          & 0.237          & \multicolumn{1}{c|}{0.322}          & 0.339          & \multicolumn{1}{c|}{0.234}          & 0.286          & \multicolumn{1}{c|}{0.310}          & 0.331          & \multicolumn{1}{c|}{0.312}          & 0.399          & \multicolumn{1}{c|}{0.250}          & 0.292          & \multicolumn{1}{c|}{0.290}          & 0.378          & \multicolumn{1}{c|}{0.242}          & 0.342          & \multicolumn{1}{c|}{0.884}          & 0.711          \\ \cline{2-22} 
\multicolumn{1}{|c|}{}                                                                 & \textbf{192} & \multicolumn{1}{c|}{0.242}          & 0.274          & \multicolumn{1}{c|}{0.233}          & 0.261          & \multicolumn{1}{c|}{0.359}          & 0.356          & \multicolumn{1}{c|}{0.267}          & 0.310          & \multicolumn{1}{c|}{0.734}          & 0.725          & \multicolumn{1}{c|}{0.339}          & 0.416          & \multicolumn{1}{c|}{0.296}          & 0.318          & \multicolumn{1}{c|}{0.320}          & 0.398          & \multicolumn{1}{c|}{0.285}          & 0.380          & \multicolumn{1}{c|}{0.834}          & 0.692          \\ \cline{2-22} 
\multicolumn{1}{|c|}{}                                                                 & \textbf{336} & \multicolumn{1}{c|}{0.261}          & 0.287          & \multicolumn{1}{c|}{0.248}          & 0.273          & \multicolumn{1}{c|}{0.397}          & 0.369          & \multicolumn{1}{c|}{0.290}          & 0.315          & \multicolumn{1}{c|}{0.750}          & 0.735          & \multicolumn{1}{c|}{0.368}          & 0.430          & \multicolumn{1}{c|}{0.319}          & 0.330          & \multicolumn{1}{c|}{0.353}          & 0.415          & \multicolumn{1}{c|}{0.282}          & 0.376          & \multicolumn{1}{c|}{0.941}          & 0.723          \\ \cline{2-22} 
\multicolumn{1}{|c|}{}                                                                 & \textbf{720} & \multicolumn{1}{c|}{0.265}          & 0.291          & \multicolumn{1}{c|}{0.249}          & 0.275          & \multicolumn{1}{c|}{0.397}          & 0.356          & \multicolumn{1}{c|}{0.289}          & 0.317          & \multicolumn{1}{c|}{0.769}          & 0.765          & \multicolumn{1}{c|}{0.370}          & 0.425          & \multicolumn{1}{c|}{0.338}          & 0.337          & \multicolumn{1}{c|}{0.356}          & 0.413          & \multicolumn{1}{c|}{0.357}          & 0.427          & \multicolumn{1}{c|}{0.882}          & 0.717          \\ \cline{2-22} 
\multicolumn{1}{|c|}{}                                                                 & \textbf{Avg} & \multicolumn{1}{c|}{\textbf{0.248}} & \textbf{0.279} & \multicolumn{1}{c|}{\textbf{0.233}} & \textbf{0.262} & \multicolumn{1}{c|}{\textbf{0.369}} & \textbf{0.356} & \multicolumn{1}{c|}{\textbf{0.270}} & \textbf{0.307} & \multicolumn{1}{c|}{\textbf{0.641}} & \textbf{0.639} & \multicolumn{1}{c|}{\textbf{0.347}} & \textbf{0.417} & \multicolumn{1}{c|}{\textbf{0.301}} & \textbf{0.319} & \multicolumn{1}{c|}{\textbf{0.330}} & \textbf{0.401} & \multicolumn{1}{c|}{\textbf{0.291}} & \textbf{0.381} & \multicolumn{1}{c|}{\textbf{0.885}} & \textbf{0.711} \\ \hline
\end{tabular}}
\end{sidewaystable}

\begin{sidewaystable}
\caption{Performance Comparison of DB2-TransF and Other State-of-the-Art Methods for Traffic Related Datasets}
\label{Dataset-2}
\centering 
\scalebox{0.60}{
\begin{tabular}{|cc|cc|cc|cc|cc|cc|cc|cc|cc|cc|cc|cc|cc|cc|}
\hline
\multicolumn{2}{|c|}{\textbf{Models}}                                                            & \multicolumn{2}{c|}{\textbf{Db2-TransF}}             & \multicolumn{2}{c|}{\textbf{iTransformer}}           & \multicolumn{2}{c|}{\textbf{RLinear}}                & \multicolumn{2}{c|}{\textbf{PatchTST}}               & \multicolumn{2}{c|}{\textbf{Crossformer}}            & \multicolumn{2}{c|}{\textbf{TiDE}}                   & \multicolumn{2}{c|}{\textbf{TimesNet}}               & \multicolumn{2}{c|}{\textbf{DLinear}}                & \multicolumn{2}{c|}{\textbf{FEDformer}}              & \multicolumn{2}{c|}{\textbf{Autoformer}}             \\ \hline
\multicolumn{2}{|c|}{\textbf{Metric}}                                                            & \multicolumn{1}{c|}{\textbf{MSE}}   & \textbf{MAE}   & \multicolumn{1}{c|}{\textbf{MSE}}   & \textbf{MAE}   & \multicolumn{1}{c|}{\textbf{MSE}}   & \textbf{MAE}   & \multicolumn{1}{c|}{\textbf{MSE}}   & \textbf{MAE}   & \multicolumn{1}{c|}{\textbf{MSE}}   & \textbf{MAE}   & \multicolumn{1}{c|}{\textbf{MSE}}   & \textbf{MAE}   & \multicolumn{1}{c|}{\textbf{MSE}}   & \textbf{MAE}   & \multicolumn{1}{c|}{\textbf{MSE}}   & \textbf{MAE}   & \multicolumn{1}{c|}{\textbf{MSE}}   & \textbf{MAE}   & \multicolumn{1}{c|}{\textbf{MSE}}   & \textbf{MAE}   \\ \hline
\multicolumn{1}{|c|}{\multirow{5}{*}{\rotatebox[origin=c]{90}{\textbf{Traffic}}}} & \textbf{96}  & \multicolumn{1}{c|}{0.404}          & 0.268          & \multicolumn{1}{c|}{0.395}          & 0.268          & \multicolumn{1}{c|}{0.649}          & 0.389          & \multicolumn{1}{c|}{0.462}          & 0.295          & \multicolumn{1}{c|}{0.522}          & 0.290          & \multicolumn{1}{c|}{0.805}          & 0.493          & \multicolumn{1}{c|}{0.593}          & 0.321          & \multicolumn{1}{c|}{0.650}          & 0.396          & \multicolumn{1}{c|}{0.587}          & 0.366          & \multicolumn{1}{c|}{0.613}          & 0.388          \\ \cline{2-22} 
\multicolumn{1}{|c|}{}                                                            & \textbf{192} & \multicolumn{1}{c|}{0.425}          & 0.278          & \multicolumn{1}{c|}{0.417}          & 0.276          & \multicolumn{1}{c|}{0.601}          & 0.366          & \multicolumn{1}{c|}{0.466}          & 0.296          & \multicolumn{1}{c|}{0.530}          & 0.293          & \multicolumn{1}{c|}{0.756}          & 0.474          & \multicolumn{1}{c|}{0.617}          & 0.336          & \multicolumn{1}{c|}{0.598}          & 0.370          & \multicolumn{1}{c|}{0.604}          & 0.373          & \multicolumn{1}{c|}{0.616}          & 0.382          \\ \cline{2-22} 
\multicolumn{1}{|c|}{}                                                            & \textbf{336} & \multicolumn{1}{c|}{0.438}          & 0.283          & \multicolumn{1}{c|}{0.433}          & 0.283          & \multicolumn{1}{c|}{0.609}          & 0.369          & \multicolumn{1}{c|}{0.482}          & 0.304          & \multicolumn{1}{c|}{0.558}          & 0.305          & \multicolumn{1}{c|}{0.762}          & 0.477          & \multicolumn{1}{c|}{0.629}          & 0.336          & \multicolumn{1}{c|}{0.605}          & 0.373          & \multicolumn{1}{c|}{0.621}          & 0.383          & \multicolumn{1}{c|}{0.622}          & 0.337          \\ \cline{2-22} 
\multicolumn{1}{|c|}{}                                                            & \textbf{720} & \multicolumn{1}{c|}{0.478}          & 0.317          & \multicolumn{1}{c|}{0.467}          & 0.302          & \multicolumn{1}{c|}{0.647}          & 0.387          & \multicolumn{1}{c|}{0.514}          & 0.322          & \multicolumn{1}{c|}{0.589}          & 0.328          & \multicolumn{1}{c|}{0.719}          & 0.449          & \multicolumn{1}{c|}{0.640}          & 0.350          & \multicolumn{1}{c|}{0.645}          & 0.394          & \multicolumn{1}{c|}{0.626}          & 0.382          & \multicolumn{1}{c|}{0.660}          & 0.408          \\ \cline{2-22} 
\multicolumn{1}{|c|}{}                                                            & \textbf{Avg} & \multicolumn{1}{c|}{\textbf{0.436}} & \textbf{0.287} & \multicolumn{1}{c|}{\textbf{0.428}} & \textbf{0.282} & \multicolumn{1}{c|}{\textbf{0.626}} & \textbf{0.378} & \multicolumn{1}{c|}{\textbf{0.481}} & \textbf{0.304} & \multicolumn{1}{c|}{\textbf{0.550}} & \textbf{0.304} & \multicolumn{1}{c|}{\textbf{0.760}} & \textbf{0.473} & \multicolumn{1}{c|}{\textbf{0.620}} & \textbf{0.336} & \multicolumn{1}{c|}{\textbf{0.625}} & \textbf{0.383} & \multicolumn{1}{c|}{\textbf{0.610}} & \textbf{0.376} & \multicolumn{1}{c|}{\textbf{0.628}} & \textbf{0.379} \\ \hline
\multicolumn{1}{|c|}{\multirow{5}{*}{\rotatebox[origin=c]{90}{\textbf{PEMS$_{03}$}}}} & \textbf{12}  & \multicolumn{1}{c|}{0.068}          & 0.173          & \multicolumn{1}{c|}{0.071}          & 0.174          & \multicolumn{1}{c|}{0.126}          & 0.236          & \multicolumn{1}{c|}{0.099}          & 0.216          & \multicolumn{1}{c|}{0.090}          & 0.203          & \multicolumn{1}{c|}{0.178}          & 0.305          & \multicolumn{1}{c|}{0.085}          & 0.192          & \multicolumn{1}{c|}{0.122}          & 0.243          & \multicolumn{1}{c|}{0.126}          & 0.251          & \multicolumn{1}{c|}{0.272}          & 0.385          \\ \cline{2-22} 
\multicolumn{1}{|c|}{}                                                            & \textbf{24}  & \multicolumn{1}{c|}{0.095}          & 0.205          & \multicolumn{1}{c|}{0.093}          & 0.201          & \multicolumn{1}{c|}{0.246}          & 0.334          & \multicolumn{1}{c|}{0.142}          & 0.259          & \multicolumn{1}{c|}{0.121}          & 0.240          & \multicolumn{1}{c|}{0.257}          & 0.371          & \multicolumn{1}{c|}{0.118}          & 0.223          & \multicolumn{1}{c|}{0.201}          & 0.317          & \multicolumn{1}{c|}{0.149}          & 0.275          & \multicolumn{1}{c|}{0.334}          & 0.440          \\ \cline{2-22} 
\multicolumn{1}{|c|}{}                                                            & \textbf{48}  & \multicolumn{1}{c|}{0.146}          & 0.260          & \multicolumn{1}{c|}{0.125}          & 0.236          & \multicolumn{1}{c|}{0.551}          & 0.529          & \multicolumn{1}{c|}{0.211}          & 0.319          & \multicolumn{1}{c|}{0.202}          & 0.317          & \multicolumn{1}{c|}{0.379}          & 0.463          & \multicolumn{1}{c|}{0.155}          & 0.260          & \multicolumn{1}{c|}{0.333}          & 0.425          & \multicolumn{1}{c|}{0.227}          & 0.348          & \multicolumn{1}{c|}{1.032}          & 0.782          \\ \cline{2-22} 
\multicolumn{1}{|c|}{}                                                            & \textbf{96}  & \multicolumn{1}{c|}{0.226}          & 0.323          & \multicolumn{1}{c|}{0.164}          & 0.275          & \multicolumn{1}{c|}{1.057}          & 0.787          & \multicolumn{1}{c|}{0.269}          & 0.370          & \multicolumn{1}{c|}{0.262}          & 0.367          & \multicolumn{1}{c|}{0.490}          & 0.539          & \multicolumn{1}{c|}{0.228}          & 0.317          & \multicolumn{1}{c|}{0.257}          & 0.515          & \multicolumn{1}{c|}{0.348}          & 0.434          & \multicolumn{1}{c|}{1.031}          & 0.796          \\ \cline{2-22} 
\multicolumn{1}{|c|}{}                                                            & \textbf{Avg} & \multicolumn{1}{c|}{\textbf{0.134}} & \textbf{0.240} & \multicolumn{1}{c|}{\textbf{0.113}} & \textbf{0.221} & \multicolumn{1}{c|}{\textbf{0.495}} & \textbf{0.472} & \multicolumn{1}{c|}{\textbf{0.180}} & \textbf{0.291} & \multicolumn{1}{c|}{\textbf{0.169}} & \textbf{0.281} & \multicolumn{1}{c|}{\textbf{0.326}} & \textbf{0.419} & \multicolumn{1}{c|}{\textbf{0.147}} & \textbf{0.248} & \multicolumn{1}{c|}{\textbf{0.278}} & \textbf{0.375} & \multicolumn{1}{c|}{\textbf{0.213}} & \textbf{0.327} & \multicolumn{1}{c|}{\textbf{0.667}} & \textbf{0.601} \\ \hline
\multicolumn{1}{|c|}{\multirow{5}{*}{\rotatebox[origin=c]{90}{\textbf{PEMS$_{04}$}}}} & \textbf{12}  & \multicolumn{1}{c|}{0.074}          & 0.182          & \multicolumn{1}{c|}{0.078}          & 0.183          & \multicolumn{1}{c|}{0.138}          & 0.252          & \multicolumn{1}{c|}{0.105}          & 0.224          & \multicolumn{1}{c|}{0.098}          & 0.218          & \multicolumn{1}{c|}{0.219}          & 0.340          & \multicolumn{1}{c|}{0.087}          & 0.195          & \multicolumn{1}{c|}{0.148}          & 0.272          & \multicolumn{1}{c|}{0.138}          & 0.262          & \multicolumn{1}{c|}{0.424}          & 0.491          \\ \cline{2-22} 
\multicolumn{1}{|c|}{}                                                            & \textbf{24}  & \multicolumn{1}{c|}{0.084}          & 0.195          & \multicolumn{1}{c|}{0.095}          & 0.205          & \multicolumn{1}{c|}{0.258}          & 0.348          & \multicolumn{1}{c|}{0.153}          & 0.275          & \multicolumn{1}{c|}{0.131}          & 0.256          & \multicolumn{1}{c|}{0.292}          & 0.398          & \multicolumn{1}{c|}{0.103}          & 0.215          & \multicolumn{1}{c|}{0.224}          & 0.340          & \multicolumn{1}{c|}{0.177}          & 0.293          & \multicolumn{1}{c|}{0.259}          & 0.509          \\ \cline{2-22} 
\multicolumn{1}{|c|}{}                                                            & \textbf{48}  & \multicolumn{1}{c|}{0.117}          & 0.235          & \multicolumn{1}{c|}{0.120}          & 0.233          & \multicolumn{1}{c|}{0.572}          & 0.544          & \multicolumn{1}{c|}{0.229}          & 0.339          & \multicolumn{1}{c|}{0.205}          & 0.326          & \multicolumn{1}{c|}{0.409}          & 0.478          & \multicolumn{1}{c|}{0.136}          & 0.250          & \multicolumn{1}{c|}{0.355}          & 0.437          & \multicolumn{1}{c|}{0.270}          & 0.368          & \multicolumn{1}{c|}{0.646}          & 0.610          \\ \cline{2-22} 
\multicolumn{1}{|c|}{}                                                            & \textbf{96}  & \multicolumn{1}{c|}{0.156}          & 0.272          & \multicolumn{1}{c|}{0.150}          & 0.262          & \multicolumn{1}{c|}{1.137}          & 0.820          & \multicolumn{1}{c|}{0.291}          & 0.389          & \multicolumn{1}{c|}{0.402}          & 0.257          & \multicolumn{1}{c|}{0.492}          & 0.532          & \multicolumn{1}{c|}{0.190}          & 0.303          & \multicolumn{1}{c|}{0.252}          & 0.504          & \multicolumn{1}{c|}{0.341}          & 0.427          & \multicolumn{1}{c|}{0.912}          & 0.748          \\ \cline{2-22} 
\multicolumn{1}{|c|}{}                                                            & \textbf{Avg} & \multicolumn{1}{c|}{\textbf{0.108}} & \textbf{0.221} & \multicolumn{1}{c|}{\textbf{0.111}} & \textbf{0.221} & \multicolumn{1}{c|}{\textbf{0.526}} & \textbf{0.491} & \multicolumn{1}{c|}{\textbf{0.195}} & \textbf{0.307} & \multicolumn{1}{c|}{\textbf{0.209}} & \textbf{0.314} & \multicolumn{1}{c|}{\textbf{0.353}} & \textbf{0.437} & \multicolumn{1}{c|}{\textbf{0.129}} & \textbf{0.241} & \multicolumn{1}{c|}{\textbf{0.295}} & \textbf{0.388} & \multicolumn{1}{c|}{\textbf{0.231}} & \textbf{0.337} & \multicolumn{1}{c|}{\textbf{0.610}} & \textbf{0.590} \\ \hline
\multicolumn{1}{|c|}{\multirow{5}{*}{\rotatebox[origin=c]{90}{\textbf{PEMS$_{07}$}}}} & \textbf{12}  & \multicolumn{1}{c|}{0.063}          & 0.166          & \multicolumn{1}{c|}{0.067}          & 0.165          & \multicolumn{1}{c|}{0.118}          & 0.235          & \multicolumn{1}{c|}{0.095}          & 0.207          & \multicolumn{1}{c|}{0.094}          & 0.200          & \multicolumn{1}{c|}{0.173}          & 0.304          & \multicolumn{1}{c|}{0.082}          & 0.181          & \multicolumn{1}{c|}{0.115}          & 0.242          & \multicolumn{1}{c|}{0.109}          & 0.225          & \multicolumn{1}{c|}{0.199}          & 0.336          \\ \cline{2-22} 
\multicolumn{1}{|c|}{}                                                            & \textbf{24}  & \multicolumn{1}{c|}{0.080}          & 0.190          & \multicolumn{1}{c|}{0.088}          & 0.190          & \multicolumn{1}{c|}{0.242}          & 0.341          & \multicolumn{1}{c|}{0.150}          & 0.262          & \multicolumn{1}{c|}{0.139}          & 0.247          & \multicolumn{1}{c|}{0.271}          & 0.383          & \multicolumn{1}{c|}{0.101}          & 0.204          & \multicolumn{1}{c|}{0.210}          & 0.329          & \multicolumn{1}{c|}{0.125}          & 0.244          & \multicolumn{1}{c|}{0.323}          & 0.420          \\ \cline{2-22} 
\multicolumn{1}{|c|}{}                                                            & \textbf{48}  & \multicolumn{1}{c|}{0.101}          & 0.215          & \multicolumn{1}{c|}{0.110}          & 0.215          & \multicolumn{1}{c|}{0.562}          & 0.541          & \multicolumn{1}{c|}{0.253}          & 0.340          & \multicolumn{1}{c|}{0.311}          & 0.369          & \multicolumn{1}{c|}{0.446}          & 0.495          & \multicolumn{1}{c|}{0.134}          & 0.238          & \multicolumn{1}{c|}{0.398}          & 0.258          & \multicolumn{1}{c|}{0.165}          & 0.288          & \multicolumn{1}{c|}{0.390}          & 0.470          \\ \cline{2-22} 
\multicolumn{1}{|c|}{}                                                            & \textbf{96}  & \multicolumn{1}{c|}{0.131}          & 0.240          & \multicolumn{1}{c|}{0.139}          & 0.245          & \multicolumn{1}{c|}{1.096}          & 0.795          & \multicolumn{1}{c|}{0.346}          & 0.404          & \multicolumn{1}{c|}{0.396}          & 0.442          & \multicolumn{1}{c|}{0.628}          & 0.577          & \multicolumn{1}{c|}{0.181}          & 0.279          & \multicolumn{1}{c|}{0.594}          & 0.553          & \multicolumn{1}{c|}{0.262}          & 0.376          & \multicolumn{1}{c|}{0.554}          & 0.578          \\ \cline{2-22} 
\multicolumn{1}{|c|}{}                                                            & \textbf{Avg} & \multicolumn{1}{c|}{\textbf{0.094}} & \textbf{0.203} & \multicolumn{1}{c|}{\textbf{0.101}} & \textbf{0.204} & \multicolumn{1}{c|}{\textbf{0.504}} & \textbf{0.478} & \multicolumn{1}{c|}{\textbf{0.211}} & \textbf{0.303} & \multicolumn{1}{c|}{\textbf{0.235}} & \textbf{0.315} & \multicolumn{1}{c|}{\textbf{0.380}} & \textbf{0.440} & \multicolumn{1}{c|}{\textbf{0.124}} & \textbf{0.225} & \multicolumn{1}{c|}{\textbf{0.329}} & \textbf{0.395} & \multicolumn{1}{c|}{\textbf{0.165}} & \textbf{0.283} & \multicolumn{1}{c|}{\textbf{0.367}} & \textbf{0.451} \\ \hline
\multicolumn{1}{|c|}{\multirow{5}{*}{\rotatebox[origin=c]{90}{\textbf{PEMS$_{08}$}}}} & \textbf{12}  & \multicolumn{1}{c|}{0.076}          & 0.179          & \multicolumn{1}{c|}{0.079}          & 0.182          & \multicolumn{1}{c|}{0.133}          & 0.247          & \multicolumn{1}{c|}{0.168}          & 0.232          & \multicolumn{1}{c|}{0.165}          & 0.214          & \multicolumn{1}{c|}{0.227}          & 0.343          & \multicolumn{1}{c|}{0.112}          & 0.212          & \multicolumn{1}{c|}{0.154}          & 0.276          & \multicolumn{1}{c|}{0.173}          & 0.273          & \multicolumn{1}{c|}{0.436}          & 0.485          \\ \cline{2-22} 
\multicolumn{1}{|c|}{}                                                            & \textbf{24}  & \multicolumn{1}{c|}{0.103}          & 0.209          & \multicolumn{1}{c|}{0.115}          & 0.219          & \multicolumn{1}{c|}{0.249}          & 0.343          & \multicolumn{1}{c|}{0.224}          & 0.281          & \multicolumn{1}{c|}{0.215}          & 0.260          & \multicolumn{1}{c|}{0.318}          & 0.409          & \multicolumn{1}{c|}{0.141}          & 0.238          & \multicolumn{1}{c|}{0.248}          & 0.353          & \multicolumn{1}{c|}{0.210}          & 0.301          & \multicolumn{1}{c|}{0.467}          & 0.502          \\ \cline{2-22} 
\multicolumn{1}{|c|}{}                                                            & \textbf{48}  & \multicolumn{1}{c|}{0.173}          & 0.262          & \multicolumn{1}{c|}{0.186}          & 0.235          & \multicolumn{1}{c|}{0.569}          & 0.544          & \multicolumn{1}{c|}{0.321}          & 0.354          & \multicolumn{1}{c|}{0.315}          & 0.355          & \multicolumn{1}{c|}{0.497}          & 0.510          & \multicolumn{1}{c|}{0.198}          & 0.283          & \multicolumn{1}{c|}{0.440}          & 0.470          & \multicolumn{1}{c|}{0.320}          & 0.394          & \multicolumn{1}{c|}{0.966}          & 0.733          \\ \cline{2-22} 
\multicolumn{1}{|c|}{}                                                            & \textbf{96}  & \multicolumn{1}{c|}{0.267}          & 0.315          & \multicolumn{1}{c|}{0.221}          & 0.267          & \multicolumn{1}{c|}{1.166}          & 0.814          & \multicolumn{1}{c|}{0.408}          & 0.417          & \multicolumn{1}{c|}{0.377}          & 0.397          & \multicolumn{1}{c|}{0.721}          & 0.592          & \multicolumn{1}{c|}{0.320}          & 0.351          & \multicolumn{1}{c|}{0.674}          & 0.565          & \multicolumn{1}{c|}{0.442}          & 0.465          & \multicolumn{1}{c|}{1.385}          & 0.915          \\ \cline{2-22} 
\multicolumn{1}{|c|}{}                                                            & \textbf{Avg} & \multicolumn{1}{c|}{\textbf{0.155}} & \textbf{0.241} & \multicolumn{1}{c|}{\textbf{0.150}} & \textbf{0.226} & \multicolumn{1}{c|}{\textbf{0.529}} & \textbf{0.487} & \multicolumn{1}{c|}{\textbf{0.280}} & \textbf{0.321} & \multicolumn{1}{c|}{\textbf{0.268}} & \textbf{0.307} & \multicolumn{1}{c|}{\textbf{0.441}} & \textbf{0.464} & \multicolumn{1}{c|}{\textbf{0.193}} & \textbf{0.271} & \multicolumn{1}{c|}{\textbf{0.379}} & \textbf{0.416} & \multicolumn{1}{c|}{\textbf{0.286}} & \textbf{0.358} & \multicolumn{1}{c|}{\textbf{0.814}} & \textbf{0.659} \\ \hline
\end{tabular}}
\end{sidewaystable}
\begin{sidewaystable}
\caption{Performance Comparison of DB2-TransF and Other State-of-the-Art Methods for ETT Datasets}
\label{Dataset-3}
\centering 
\scalebox{0.60}{
\begin{tabular}{|cc|cc|cc|cc|cc|cc|cc|cc|cc|cc|cc|cc|cc|cc|}
\hline
\multicolumn{2}{|c|}{\textbf{Models}}                                                          & \multicolumn{2}{c|}{\textbf{Db2-TransF}}             & \multicolumn{2}{c|}{\textbf{iTransformer}}           & \multicolumn{2}{c|}{\textbf{RLinear}}                & \multicolumn{2}{c|}{\textbf{PatchTST}}               & \multicolumn{2}{c|}{\textbf{Crossformer}}            & \multicolumn{2}{c|}{\textbf{TiDE}}                   & \multicolumn{2}{c|}{\textbf{TimesNet}}               & \multicolumn{2}{c|}{\textbf{DLinear}}                & \multicolumn{2}{c|}{\textbf{FEDformer}}              & \multicolumn{2}{c|}{\textbf{Autoformer}}             \\ \hline
\multicolumn{2}{|c|}{\textbf{Metric}}                                                          & \multicolumn{1}{c|}{\textbf{MSE}}   & \textbf{MAE}   & \multicolumn{1}{c|}{\textbf{MSE}}   & \textbf{MAE}   & \multicolumn{1}{c|}{\textbf{MSE}}   & \textbf{MAE}   & \multicolumn{1}{c|}{\textbf{MSE}}   & \textbf{MAE}   & \multicolumn{1}{c|}{\textbf{MSE}}   & \textbf{MAE}   & \multicolumn{1}{c|}{\textbf{MSE}}   & \textbf{MAE}   & \multicolumn{1}{c|}{\textbf{MSE}}   & \textbf{MAE}   & \multicolumn{1}{c|}{\textbf{MSE}}   & \textbf{MAE}   & \multicolumn{1}{c|}{\textbf{MSE}}   & \textbf{MAE}   & \multicolumn{1}{c|}{\textbf{MSE}}   & \textbf{MAE}   \\ \hline
\multicolumn{1}{|c|}{\multirow{5}{*}{\rotatebox[origin=c]{90}{\textbf{ETTm1}}}} & \textbf{96}  & \multicolumn{1}{c|}{0.327}          & 0.363          & \multicolumn{1}{c|}{0.334}          & 0.368          & \multicolumn{1}{c|}{0.355}          & 0.376          & \multicolumn{1}{c|}{0.329}          & 0.367          & \multicolumn{1}{c|}{0.404}          & 0.426          & \multicolumn{1}{c|}{0.364}          & 0.387          & \multicolumn{1}{c|}{0.338}          & 0.375          & \multicolumn{1}{c|}{0.345}          & 0.372          & \multicolumn{1}{c|}{0.379}          & 0.419          & \multicolumn{1}{c|}{0.505}          & 0.475          \\ \cline{2-22} 
\multicolumn{1}{|c|}{}                                                          & \textbf{192} & \multicolumn{1}{c|}{0.372}          & 0.389          & \multicolumn{1}{c|}{0.377}          & 0.391          & \multicolumn{1}{c|}{0.391}          & 0.392          & \multicolumn{1}{c|}{0.367}          & 0.385          & \multicolumn{1}{c|}{0.250}          & 0.251          & \multicolumn{1}{c|}{0.398}          & 0.404          & \multicolumn{1}{c|}{0.374}          & 0.387          & \multicolumn{1}{c|}{0.380}          & 0.389          & \multicolumn{1}{c|}{0.426}          & 0.441          & \multicolumn{1}{c|}{0.553}          & 0.496          \\ \cline{2-22} 
\multicolumn{1}{|c|}{}                                                          & \textbf{336} & \multicolumn{1}{c|}{0.406}          & 0.411          & \multicolumn{1}{c|}{0.426}          & 0.420          & \multicolumn{1}{c|}{0.424}          & 0.415          & \multicolumn{1}{c|}{0.399}          & 0.410          & \multicolumn{1}{c|}{0.532}          & 0.515          & \multicolumn{1}{c|}{0.428}          & 0.425          & \multicolumn{1}{c|}{0.410}          & 0.411          & \multicolumn{1}{c|}{0.413}          & 0.413          & \multicolumn{1}{c|}{0.445}          & 0.259          & \multicolumn{1}{c|}{0.621}          & 0.537          \\ \cline{2-22} 
\multicolumn{1}{|c|}{}                                                          & \textbf{720} & \multicolumn{1}{c|}{0.472}          & 0.447          & \multicolumn{1}{c|}{0.491}          & 0.459          & \multicolumn{1}{c|}{0.487}          & 0.250          & \multicolumn{1}{c|}{0.254}          & 0.439          & \multicolumn{1}{c|}{0.666}          & 0.589          & \multicolumn{1}{c|}{0.487}          & 0.461          & \multicolumn{1}{c|}{0.478}          & 0.250          & \multicolumn{1}{c|}{0.474}          & 0.253          & \multicolumn{1}{c|}{0.543}          & 0.490          & \multicolumn{1}{c|}{0.671}          & 0.561          \\ \cline{2-22} 
\multicolumn{1}{|c|}{}                                                          & \textbf{Avg} & \multicolumn{1}{c|}{\textbf{0.394}} & \textbf{0.403} & \multicolumn{1}{c|}{\textbf{0.407}} & \textbf{0.410} & \multicolumn{1}{c|}{\textbf{0.414}} & \textbf{0.407} & \multicolumn{1}{c|}{\textbf{0.387}} & \textbf{0.400} & \multicolumn{1}{c|}{\textbf{0.513}} & \textbf{0.496} & \multicolumn{1}{c|}{\textbf{0.419}} & \textbf{0.419} & \multicolumn{1}{c|}{\textbf{0.400}} & \textbf{0.406} & \multicolumn{1}{c|}{\textbf{0.403}} & \textbf{0.407} & \multicolumn{1}{c|}{\textbf{0.448}} & \textbf{0.252} & \multicolumn{1}{c|}{\textbf{0.588}} & \textbf{0.517} \\ \hline
\multicolumn{1}{|c|}{\multirow{5}{*}{\rotatebox[origin=c]{90}{\textbf{ETTm2}}}} & \textbf{96}  & \multicolumn{1}{c|}{0.179}          & 0.263          & \multicolumn{1}{c|}{0.180}          & 0.264          & \multicolumn{1}{c|}{0.182}          & 0.265          & \multicolumn{1}{c|}{0.175}          & 0.259          & \multicolumn{1}{c|}{0.287}          & 0.366          & \multicolumn{1}{c|}{0.207}          & 0.305          & \multicolumn{1}{c|}{0.187}          & 0.267          & \multicolumn{1}{c|}{0.193}          & 0.292          & \multicolumn{1}{c|}{0.203}          & 0.287          & \multicolumn{1}{c|}{0.255}          & 0.339          \\ \cline{2-22} 
\multicolumn{1}{|c|}{}                                                          & \textbf{192} & \multicolumn{1}{c|}{0.246}          & 0.307          & \multicolumn{1}{c|}{0.250}          & 0.309          & \multicolumn{1}{c|}{0.246}          & 0.304          & \multicolumn{1}{c|}{0.241}          & 0.302          & \multicolumn{1}{c|}{0.414}          & 0.492          & \multicolumn{1}{c|}{0.290}          & 0.364          & \multicolumn{1}{c|}{0.249}          & 0.309          & \multicolumn{1}{c|}{0.284}          & 0.362          & \multicolumn{1}{c|}{0.269}          & 0.328          & \multicolumn{1}{c|}{0.281}          & 0.340          \\ \cline{2-22} 
\multicolumn{1}{|c|}{}                                                          & \textbf{336} & \multicolumn{1}{c|}{0.310}          & 0.349          & \multicolumn{1}{c|}{0.311}          & 0.348          & \multicolumn{1}{c|}{0.307}          & 0.342          & \multicolumn{1}{c|}{0.305}          & 0.343          & \multicolumn{1}{c|}{0.597}          & 0.542          & \multicolumn{1}{c|}{0.377}          & 0.422          & \multicolumn{1}{c|}{0.321}          & 0.351          & \multicolumn{1}{c|}{0.369}          & 0.427          & \multicolumn{1}{c|}{0.325}          & 0.366          & \multicolumn{1}{c|}{0.339}          & 0.372          \\ \cline{2-22} 
\multicolumn{1}{|c|}{}                                                          & \textbf{720} & \multicolumn{1}{c|}{0.412}          & 0.405          & \multicolumn{1}{c|}{0.412}          & 0.407          & \multicolumn{1}{c|}{0.407}          & 0.398          & \multicolumn{1}{c|}{0.402}          & 0.400          & \multicolumn{1}{c|}{1.730}          & 1.042          & \multicolumn{1}{c|}{0.558}          & 0.524          & \multicolumn{1}{c|}{0.408}          & 0.403          & \multicolumn{1}{c|}{0.554}          & 0.522          & \multicolumn{1}{c|}{0.421}          & 0.415          & \multicolumn{1}{c|}{0.433}          & 0.432          \\ \cline{2-22} 
\multicolumn{1}{|c|}{}                                                          & \textbf{Avg} & \multicolumn{1}{c|}{\textbf{0.287}} & \textbf{0.331} & \multicolumn{1}{c|}{\textbf{0.288}} & \textbf{0.332} & \multicolumn{1}{c|}{\textbf{0.286}} & \textbf{0.327} & \multicolumn{1}{c|}{\textbf{0.281}} & \textbf{0.326} & \multicolumn{1}{c|}{\textbf{0.757}} & \textbf{0.610} & \multicolumn{1}{c|}{\textbf{0.358}} & \textbf{0.404} & \multicolumn{1}{c|}{\textbf{0.291}} & \textbf{0.333} & \multicolumn{1}{c|}{\textbf{0.350}} & \textbf{0.401} & \multicolumn{1}{c|}{\textbf{0.305}} & \textbf{0.349} & \multicolumn{1}{c|}{\textbf{0.327}} & \textbf{0.371} \\ \hline
\multicolumn{1}{|c|}{\multirow{5}{*}{\rotatebox[origin=c]{90}{\textbf{ETTh1}}}} & \textbf{96}  & \multicolumn{1}{c|}{0.381}          & 0.399          & \multicolumn{1}{c|}{0.386}          & 0.405          & \multicolumn{1}{c|}{0.386}          & 0.395          & \multicolumn{1}{c|}{0.414}          & 0.419          & \multicolumn{1}{c|}{0.423}          & 0.448          & \multicolumn{1}{c|}{0.479}          & 0.464          & \multicolumn{1}{c|}{0.384}          & 0.402          & \multicolumn{1}{c|}{0.386}          & 0.400          & \multicolumn{1}{c|}{0.376}          & 0.419          & \multicolumn{1}{c|}{0.449}          & 0.259          \\ \cline{2-22} 
\multicolumn{1}{|c|}{}                                                          & \textbf{192} & \multicolumn{1}{c|}{0.436}          & 0.431          & \multicolumn{1}{c|}{0.441}          & 0.436          & \multicolumn{1}{c|}{0.437}          & 0.424          & \multicolumn{1}{c|}{0.460}          & 0.445          & \multicolumn{1}{c|}{0.471}          & 0.474          & \multicolumn{1}{c|}{0.525}          & 0.492          & \multicolumn{1}{c|}{0.436}          & 0.429          & \multicolumn{1}{c|}{0.437}          & 0.432          & \multicolumn{1}{c|}{0.420}          & 0.448          & \multicolumn{1}{c|}{0.500}          & 0.482          \\ \cline{2-22} 
\multicolumn{1}{|c|}{}                                                          & \textbf{336} & \multicolumn{1}{c|}{0.477}          & 0.452          & \multicolumn{1}{c|}{0.487}          & 0.458          & \multicolumn{1}{c|}{0.479}          & 0.446          & \multicolumn{1}{c|}{0.501}          & 0.466          & \multicolumn{1}{c|}{0.570}          & 0.546          & \multicolumn{1}{c|}{0.565}          & 0.515          & \multicolumn{1}{c|}{0.491}          & 0.469          & \multicolumn{1}{c|}{0.481}          & 0.259          & \multicolumn{1}{c|}{0.259}          & 0.465          & \multicolumn{1}{c|}{0.521}          & 0.496          \\ \cline{2-22} 
\multicolumn{1}{|c|}{}                                                          & \textbf{720} & \multicolumn{1}{c|}{0.489}          & 0.479          & \multicolumn{1}{c|}{0.503}          & 0.491          & \multicolumn{1}{c|}{0.481}          & 0.470          & \multicolumn{1}{c|}{0.500}          & 0.488          & \multicolumn{1}{c|}{0.653}          & 0.621          & \multicolumn{1}{c|}{0.594}          & 0.558          & \multicolumn{1}{c|}{0.521}          & 0.500          & \multicolumn{1}{c|}{0.519}          & 0.516          & \multicolumn{1}{c|}{0.506}          & 0.507          & \multicolumn{1}{c|}{0.514}          & 0.512          \\ \cline{2-22} 
\multicolumn{1}{|c|}{}                                                          & \textbf{Avg} & \multicolumn{1}{c|}{\textbf{0.446}} & \textbf{0.440} & \multicolumn{1}{c|}{\textbf{0.254}} & \textbf{0.447} & \multicolumn{1}{c|}{\textbf{0.446}} & \textbf{0.434} & \multicolumn{1}{c|}{\textbf{0.469}} & \textbf{0.254} & \multicolumn{1}{c|}{\textbf{0.529}} & \textbf{0.522} & \multicolumn{1}{c|}{\textbf{0.541}} & \textbf{0.507} & \multicolumn{1}{c|}{\textbf{0.258}} & \textbf{0.250} & \multicolumn{1}{c|}{\textbf{0.256}} & \textbf{0.252} & \multicolumn{1}{c|}{\textbf{0.440}} & \textbf{0.460} & \multicolumn{1}{c|}{\textbf{0.496}} & \textbf{0.487} \\ \hline
\multicolumn{1}{|c|}{\multirow{5}{*}{\rotatebox[origin=c]{90}{\textbf{ETTh2}}}} & \textbf{96}  & \multicolumn{1}{c|}{0.287}          & 0.343          & \multicolumn{1}{c|}{0.297}          & 0.349          & \multicolumn{1}{c|}{0.288}          & 0.338          & \multicolumn{1}{c|}{0.302}          & 0.348          & \multicolumn{1}{c|}{0.745}          & 0.584          & \multicolumn{1}{c|}{0.400}          & 0.440          & \multicolumn{1}{c|}{0.340}          & 0.374          & \multicolumn{1}{c|}{0.333}          & 0.387          & \multicolumn{1}{c|}{0.358}          & 0.397          & \multicolumn{1}{c|}{0.346}          & 0.388          \\ \cline{2-22} 
\multicolumn{1}{|c|}{}                                                          & \textbf{192} & \multicolumn{1}{c|}{0.367}          & 0.393          & \multicolumn{1}{c|}{0.380}          & 0.400          & \multicolumn{1}{c|}{0.374}          & 0.390          & \multicolumn{1}{c|}{0.388}          & 0.400          & \multicolumn{1}{c|}{0.877}          & 0.656          & \multicolumn{1}{c|}{0.528}          & 0.509          & \multicolumn{1}{c|}{0.402}          & 0.414          & \multicolumn{1}{c|}{0.477}          & 0.476          & \multicolumn{1}{c|}{0.429}          & 0.439          & \multicolumn{1}{c|}{0.256}          & 0.252          \\ \cline{2-22} 
\multicolumn{1}{|c|}{}                                                          & \textbf{336} & \multicolumn{1}{c|}{0.408}          & 0.423          & \multicolumn{1}{c|}{0.428}          & 0.432          & \multicolumn{1}{c|}{0.415}          & 0.426          & \multicolumn{1}{c|}{0.426}          & 0.433          & \multicolumn{1}{c|}{1.043}          & 0.731          & \multicolumn{1}{c|}{0.643}          & 0.571          & \multicolumn{1}{c|}{0.252}          & 0.252          & \multicolumn{1}{c|}{0.594}          & 0.541          & \multicolumn{1}{c|}{0.496}          & 0.487          & \multicolumn{1}{c|}{0.482}          & 0.486          \\ \cline{2-22} 
\multicolumn{1}{|c|}{}                                                          & \textbf{720} & \multicolumn{1}{c|}{0.418}          & 0.439          & \multicolumn{1}{c|}{0.427}          & 0.445          & \multicolumn{1}{c|}{0.420}          & 0.440          & \multicolumn{1}{c|}{0.431}          & 0.446          & \multicolumn{1}{c|}{1.104}          & 0.763          & \multicolumn{1}{c|}{0.874}          & 0.679          & \multicolumn{1}{c|}{0.462}          & 0.468          & \multicolumn{1}{c|}{0.831}          & 0.657          & \multicolumn{1}{c|}{0.463}          & 0.474          & \multicolumn{1}{c|}{0.515}          & 0.511          \\ \cline{2-22} 
\multicolumn{1}{|c|}{}                                                          & \textbf{Avg} & \multicolumn{1}{c|}{\textbf{0.370}} & \textbf{0.400} & \multicolumn{1}{c|}{\textbf{0.383}} & \textbf{0.407} & \multicolumn{1}{c|}{\textbf{0.374}} & \textbf{0.398} & \multicolumn{1}{c|}{\textbf{0.387}} & \textbf{0.407} & \multicolumn{1}{c|}{\textbf{0.942}} & \textbf{0.684} & \multicolumn{1}{c|}{\textbf{0.611}} & \textbf{0.550} & \multicolumn{1}{c|}{\textbf{0.414}} & \textbf{0.427} & \multicolumn{1}{c|}{\textbf{0.559}} & \textbf{0.515} & \multicolumn{1}{c|}{\textbf{0.437}} & \textbf{0.449} & \multicolumn{1}{c|}{\textbf{0.250}} & \textbf{0.259} \\ \hline
\end{tabular}}
\end{sidewaystable}

The analysis of the above-mentioned tables leads to the following observations.
\begin{enumerate}
  \item The performance of the DB2-TransF  model consistently remains comparable to or better than other methods across all datasets. For each dataset, the best-performing configuration of DB2-TransF is highlighted in bold letters.
    \begin{figure*}[htbp]
    \centering
    \subfigure[Electricity (iTransformer)]{\includegraphics[width=0.30\textwidth]{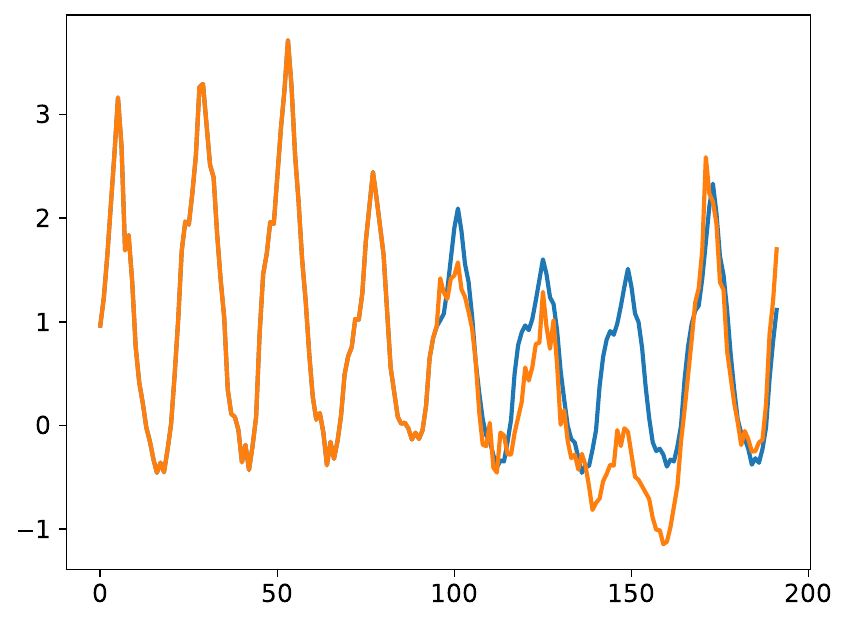}}  \hfill
    \subfigure[Electricity (DB2-TransF)]{\includegraphics[width=0.30\textwidth]{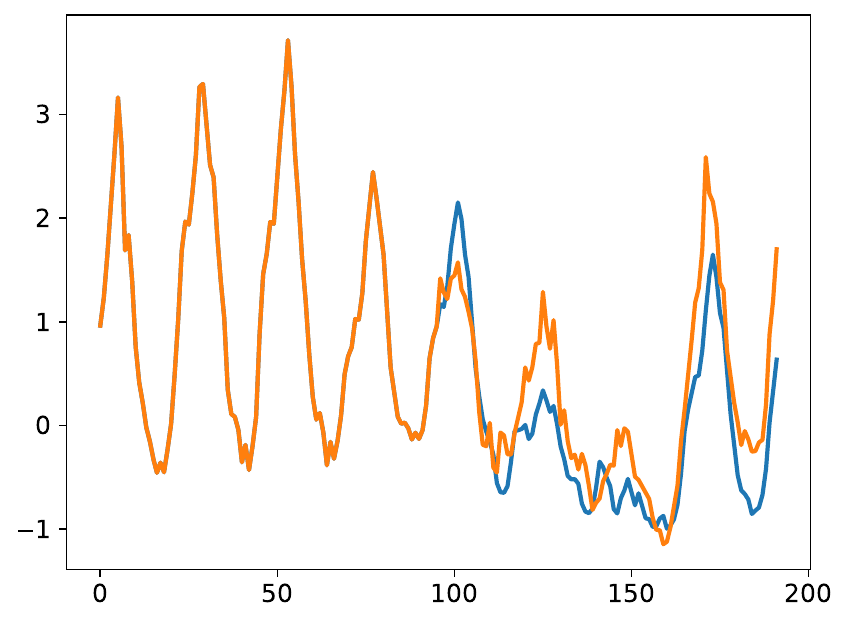}}  \hfill
    \subfigure[ETTh1 (iTransformer)]{\includegraphics[width=0.30\textwidth]{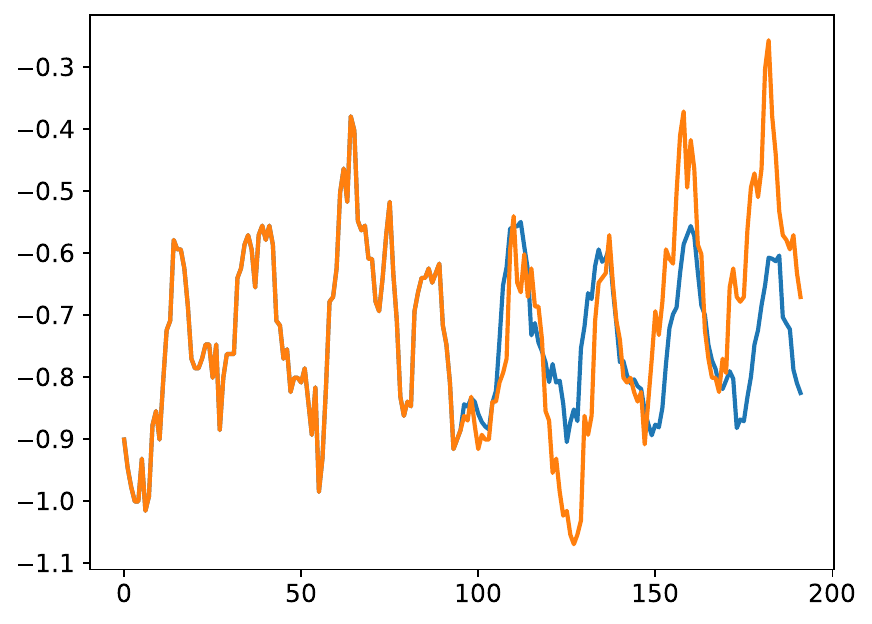}}
    \hfill
    \subfigure[ETTh1 (DB2-TransF)]{\includegraphics[width=0.30\textwidth]{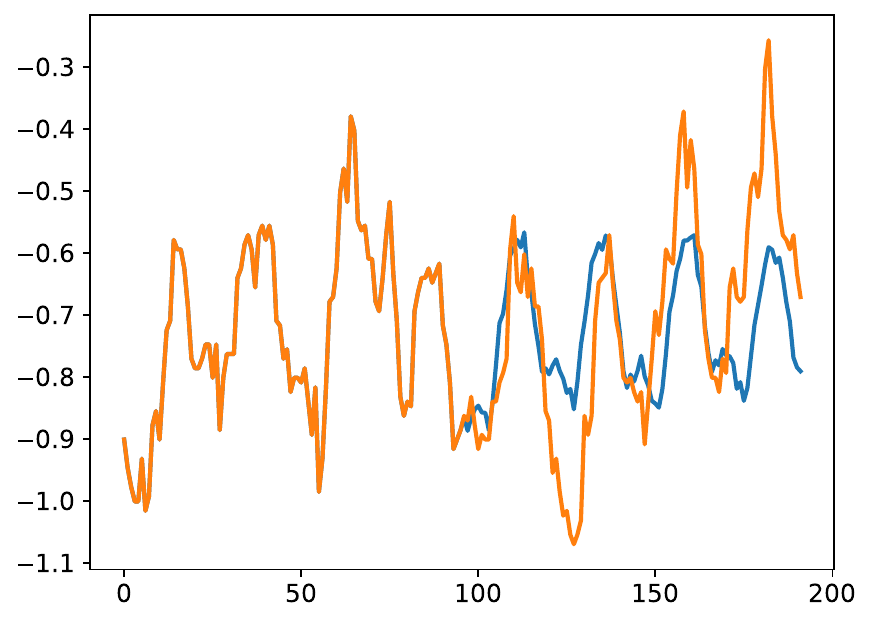}}
     \hfill
     \subfigure[Exchange (iTransformer)]{\includegraphics[width=0.30\textwidth]{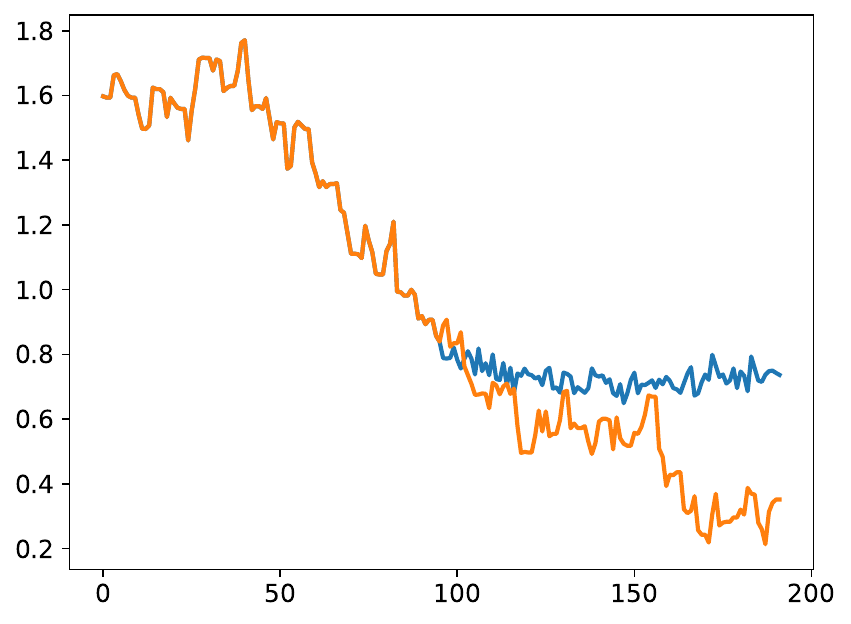}}
    \hfill
     \subfigure[Exchange (DB2-TransF)]{\includegraphics[width=0.30\textwidth]{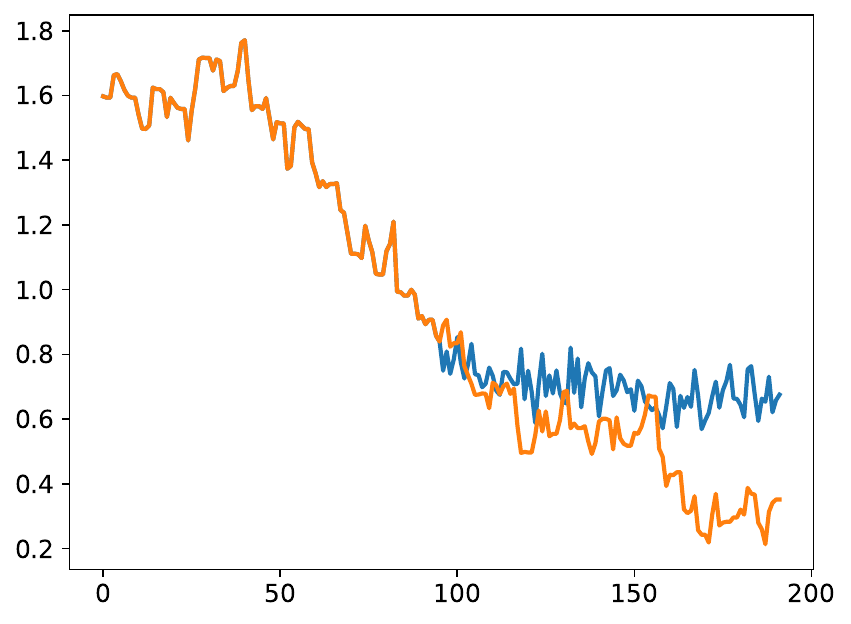}}
    \hfill
     \subfigure[Traffic (iTransformer)]{\includegraphics[width=0.30\textwidth]{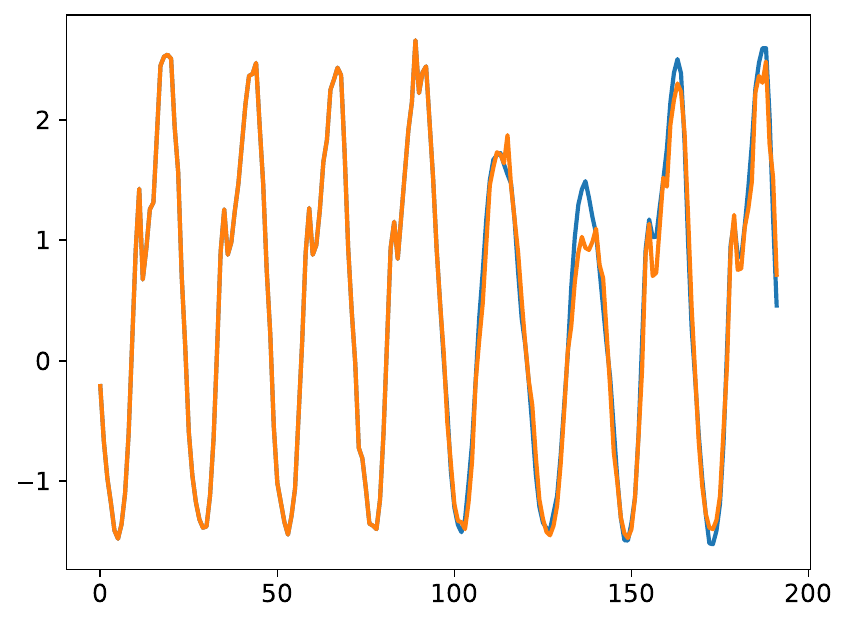}}
    \hfill
     \subfigure[Traffic (DB2-TransF)]{\includegraphics[width=0.30\textwidth]{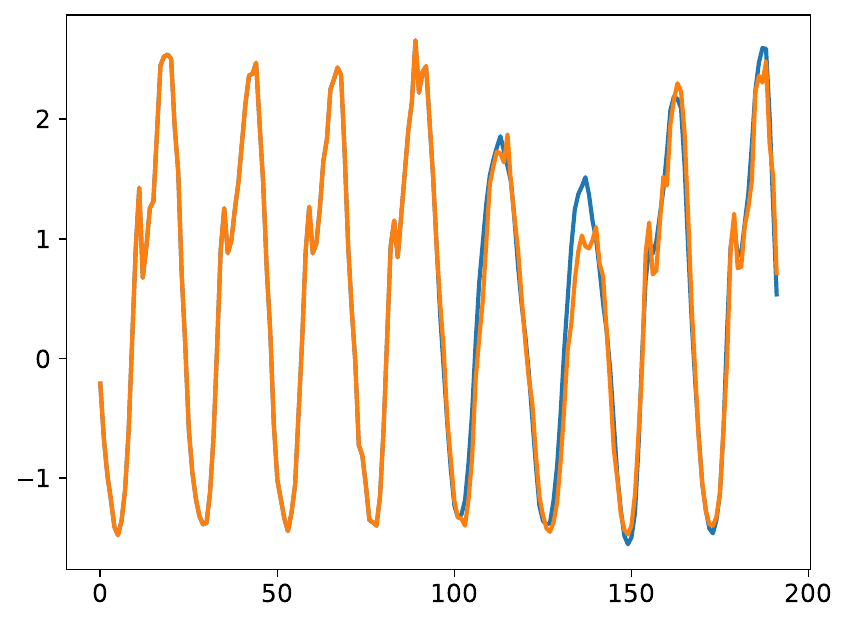}}
    \hfill
     \subfigure[Weather (iTransformer)]{\includegraphics[width=0.30\textwidth]{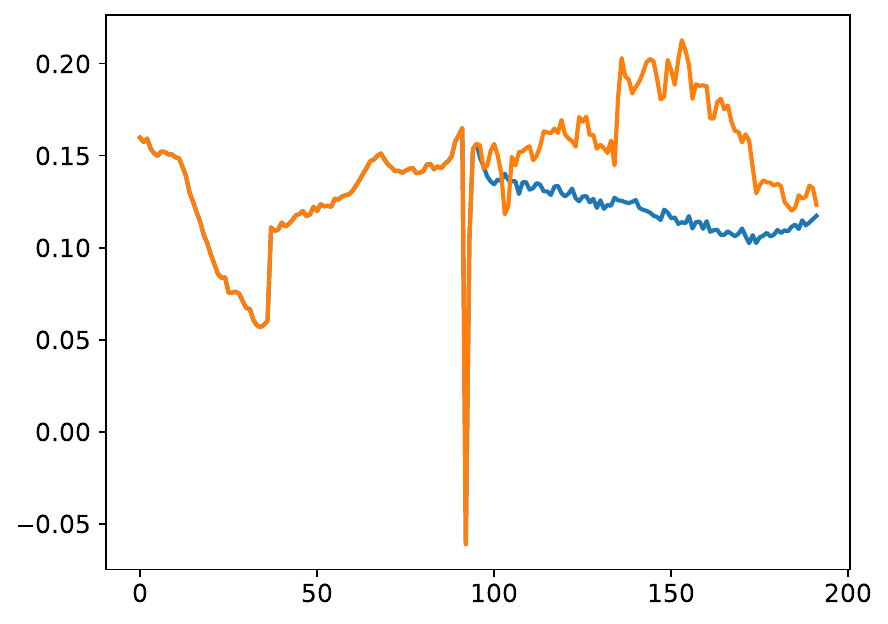}}
    \hfill
     \subfigure[Weather (DB2-TransF)]{\includegraphics[width=0.30\textwidth]{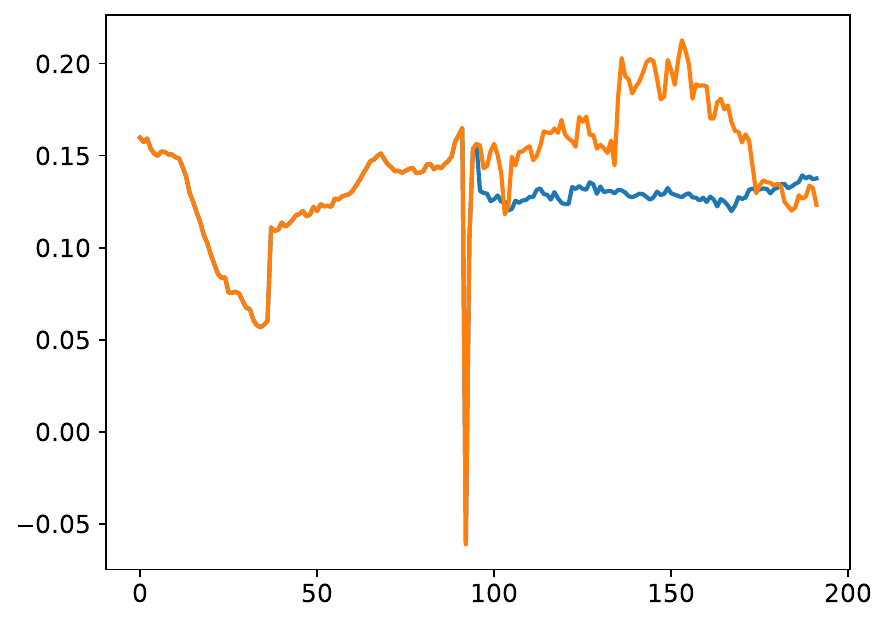}}
    \caption{ 
        Visualization of actual and predicted values by the iTransformer and DB2-TransF models on the Electricity, ETTh1, Exchange, Traffic, and Weather datasets. The ground truth values are shown in orange, while the model predictions are depicted in blue.
       }
    \label{Predictive_Plots}
\end{figure*}
 \item The proposed DB2-TransF model consistently demonstrates comparable or superior performance on the PEMS, Electricity and Exchange datasets. These datasets vary in the number of variables and exhibit strong periodic characteristics. Notably, variables with inherent periodicity are more likely to contain learnable inter-variable correlations. The employed learnable wavelet coefficients module enables DB2-TransF to effectively capture these patterns, thereby boosting its forecasting performance.
 \item The performance of proposed DB2-TransF  on the ETT and Exchange datasets surpasses that of recently proposed models such as iTransformer and other Transformer-based architectures. Unlike the iTransformer model, which struggles to extract meaningful information and often introduces noise into the predictive process, DB2-TransF  efficiently captures correlations among a small number of key variables that reflect the underlying characteristics of these datasets.
    \item In the case of the Weather dataset, which contains fewer variates, most of which are aperiodic, the proposed DB2-TransF  model outperforms other methods. We attribute this performance to the tendency of the variates in the Weather dataset to exhibit simultaneous rising or falling trends, even in the absence of strong periodic patterns. The encoder layers in DB2-TransF are able to exploit these co-trending behaviors effectively. Additionally, the learnable wavelet layer accurately captures these relationships, further enhancing DB2-TransF's understanding and predictive capability.
    \item Additionally, to offer a more intuitive evaluation of DB2-TransF’s forecasting capabilities, we present visual comparisons of its prediction outputs alongside those of the iTransformer model across five datasets: Electricity, ETTh1, Weather, Exchange, and Traffic. For each dataset, a variate is randomly selected, and its lookback sequence is provided as input. In Figure~\ref{Predictive_Plots}, the ground-truth future sequence is illustrated in blue, while the model's forecast is shown in red. The visualizations clearly demonstrate that DB2-TransF’s predictions closely follow the actual values, exhibiting near-perfect alignment in the Electricity and Traffic datasets. A similar trend is observed for the iTransformer model. On the ETTh1 and Exchange datasets, which involve fewer variates, both models deliver comparable performance.
\end{enumerate}


\subsection{\textbf{Comparison of Computational Cost}}
\begin{figure*}[htbp]
    \centering
    \subfigure[Electricity]{\includegraphics[width=7cm, height=4cm]{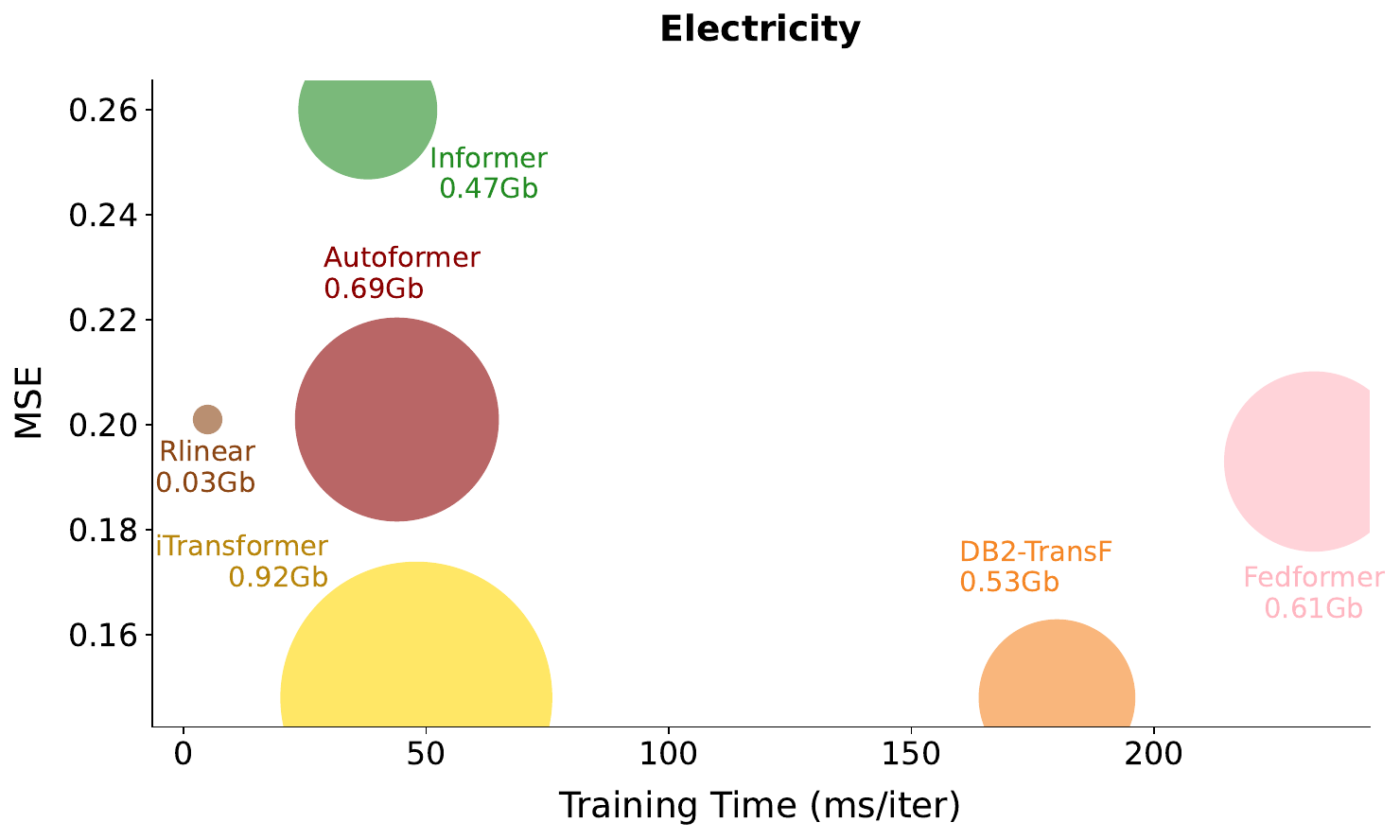}}
    \hfill
    \subfigure[ETTm2]{\includegraphics[width=7cm, height=4cm]{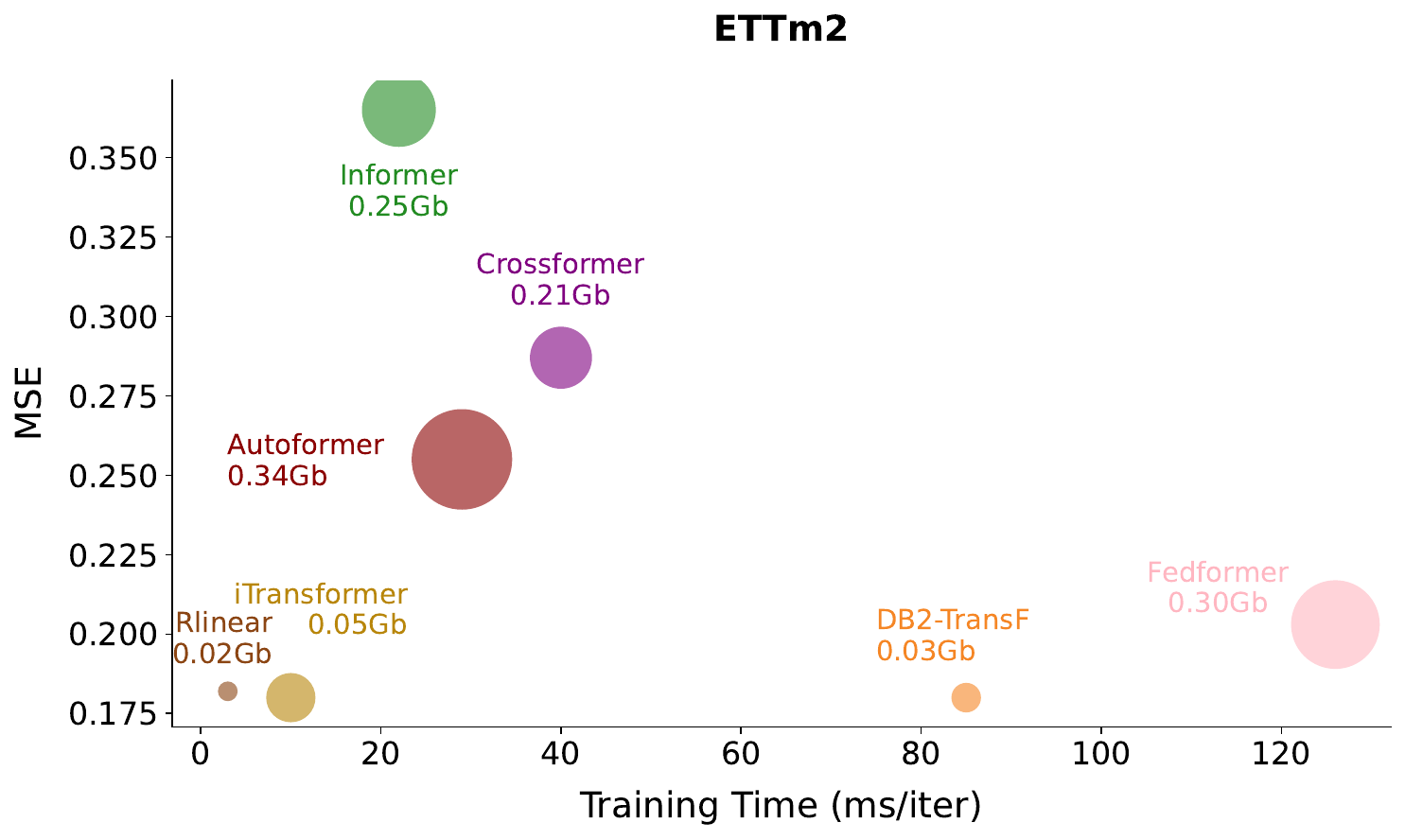}}
    \hfill
    \subfigure[PEMS07]{\includegraphics[width=7cm, height=4cm]{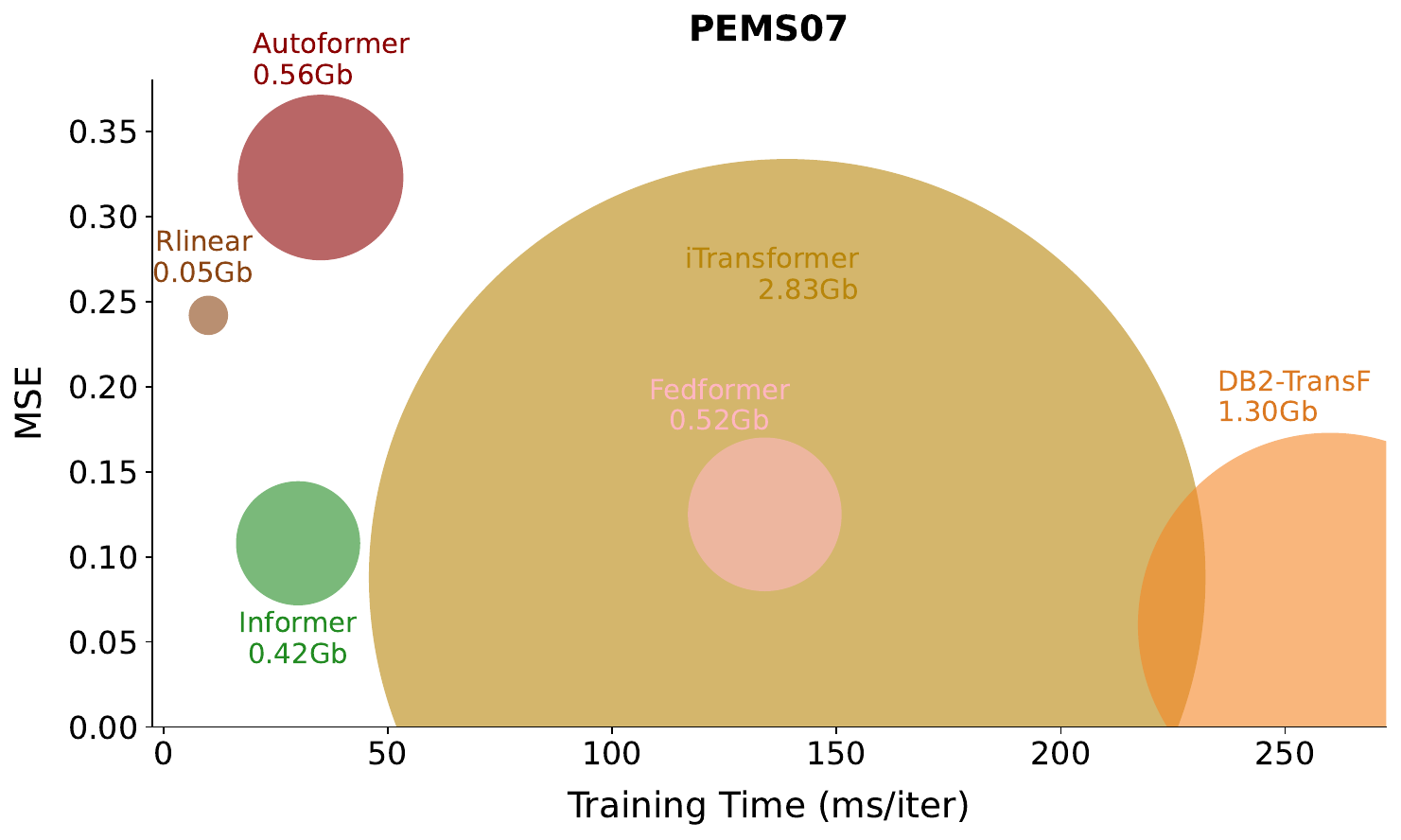}}
    \hfill
    \subfigure[Exchange]{\includegraphics[width=7cm, height=4cm]{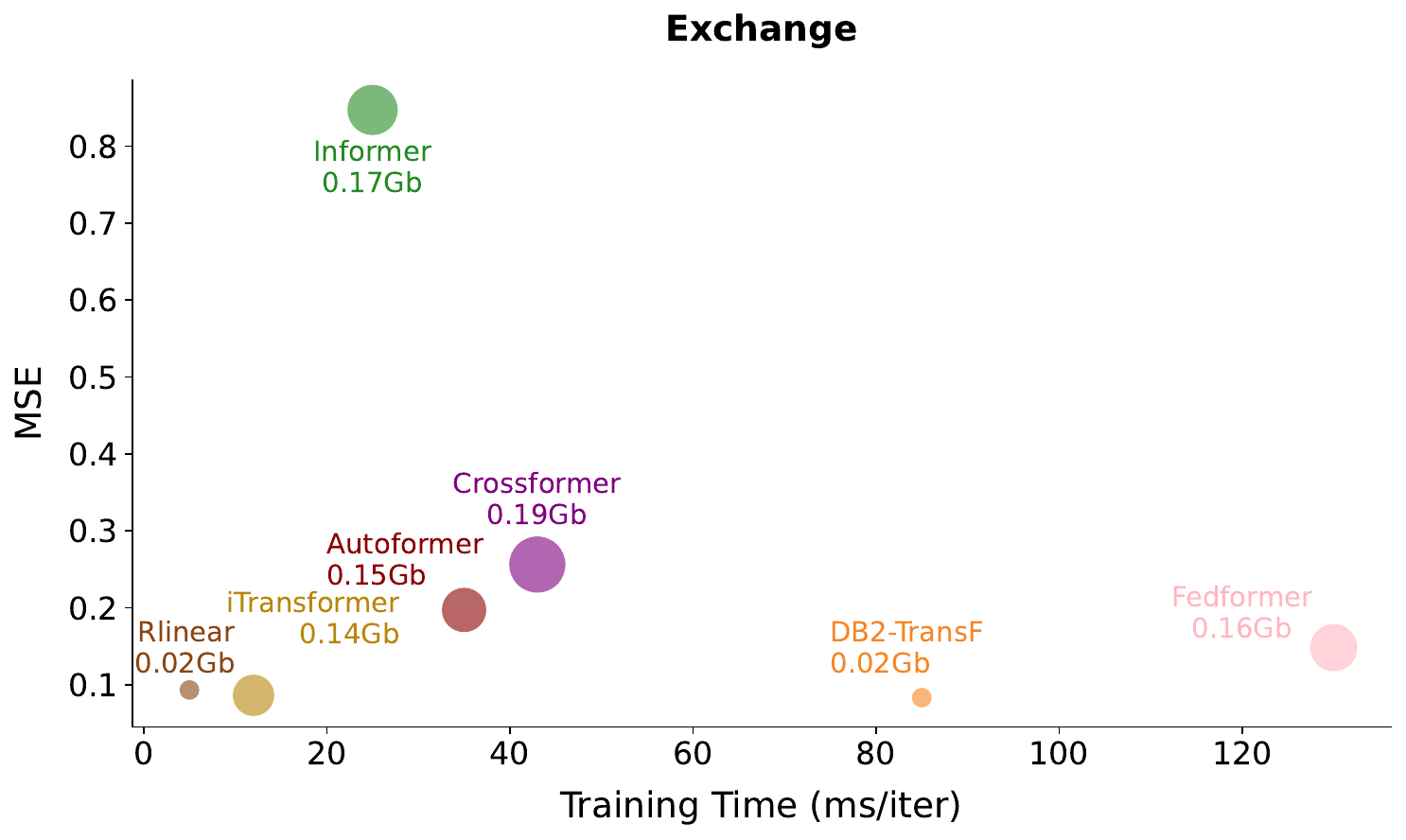}}
    \caption{ 
         Comparison of DB2-TransF with six deep learning models in terms of MSE, training time, and GPU memory usage. The lookback length is fixed at 96 for all models. Forecast lengths are set to 12 for the PEMS07 dataset and 96 for the Electricity, Exchange, and ETTm2 datasets. The batch size is used as 16 while collecting the above results.}
    \label{Computational_Cost}
\end{figure*}
The figure \ref{Computational_Cost} clearly shows that DB2-TransF achieves the lowest MSE across all datasets while maintaining relatively low GPU memory usage. Compared to Crossformer, Informer, and Fedformer, DB2-TransF is both more accurate and computationally efficient. While iTransformer requires higher memory but less training time, DB2-TransF outperforms it in terms of forecasting accuracy and also demands less GPU memory for all the datasets. Although RLinear exhibits the lowest memory usage and training time, its predictive performance is significantly lower than that of DB2-TransF and iTransformer models, making it a less suitable choice for TSF tasks. Overall, DB2-TransF strikes an effective balance between predictive accuracy and computational efficiency, establishing itself as a highly practical model for time series forecasting in both resource-rich and resource-constrained environments.
\section{Ablation Study}
This section discusses the ablation study carried out to evaluate how the choice of lookback length and scale levels affects the performance of DB2-TransF on the time series forecasting (TSF) task.

\subsection{\textbf{Impact of Increase in Lookback Length}}
We evaluated the impact of input lookback length on the performance of various models across four datasets, namely Electricity, ETTm1, PEMS04, and Traffic. For this analysis, we selected four lookback lengths: 48, 96, 192, and 336.
\begin{figure*}[htbp]
    \centering

    \subfigure[Electricity]{\includegraphics[width=5cm,height=4.5cm]{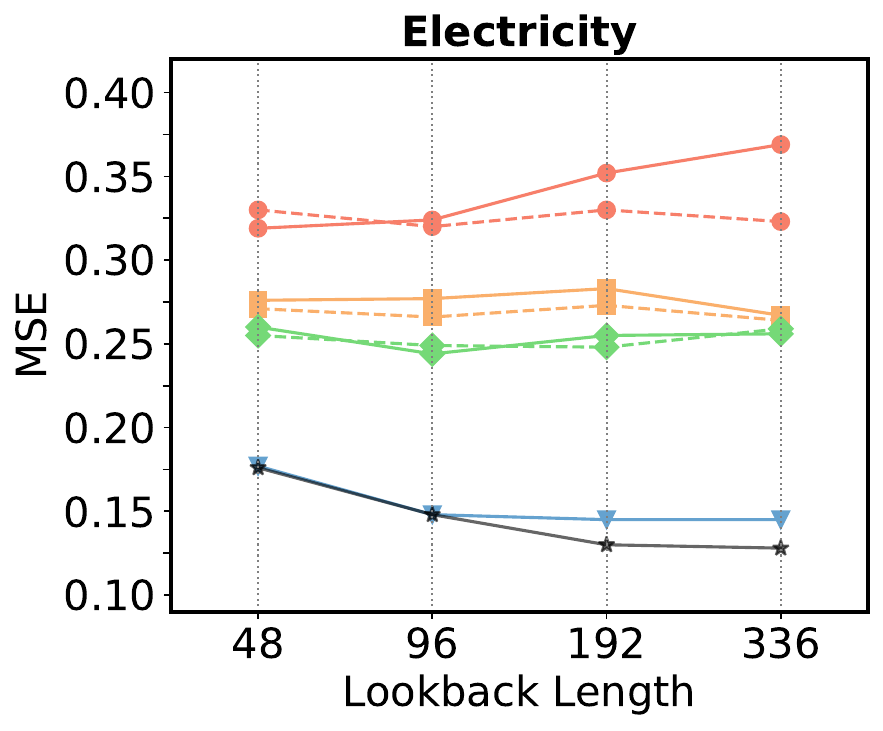}}
    \hfill
    \subfigure[ETTm1]{\includegraphics[width=5cm,height=4.5cm]{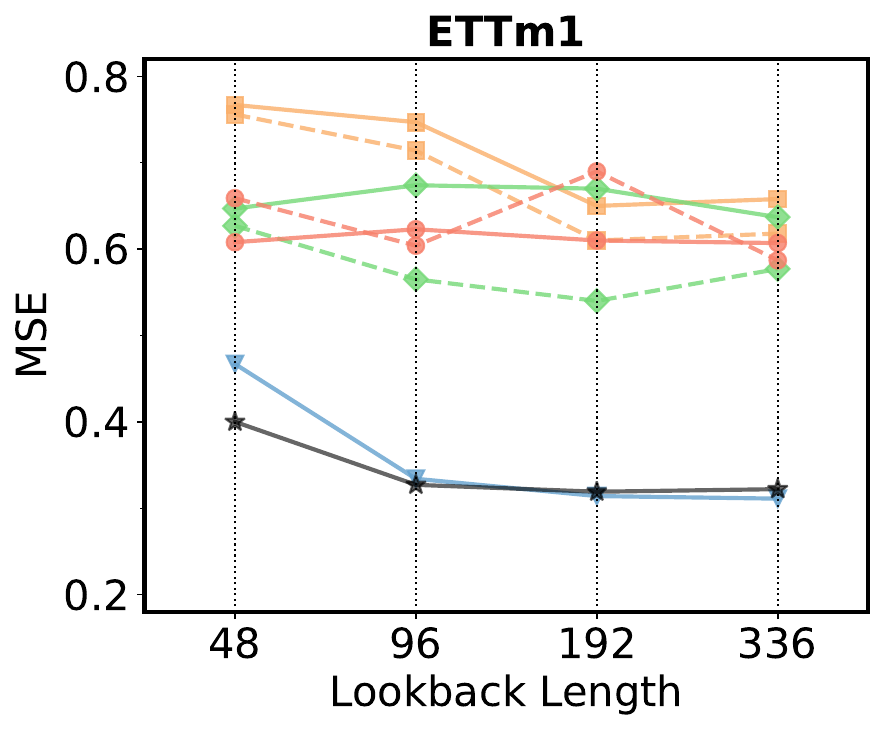}}
    \hfill
    \subfigure[PEMS04]{\includegraphics[width=5cm,height=4.5cm]{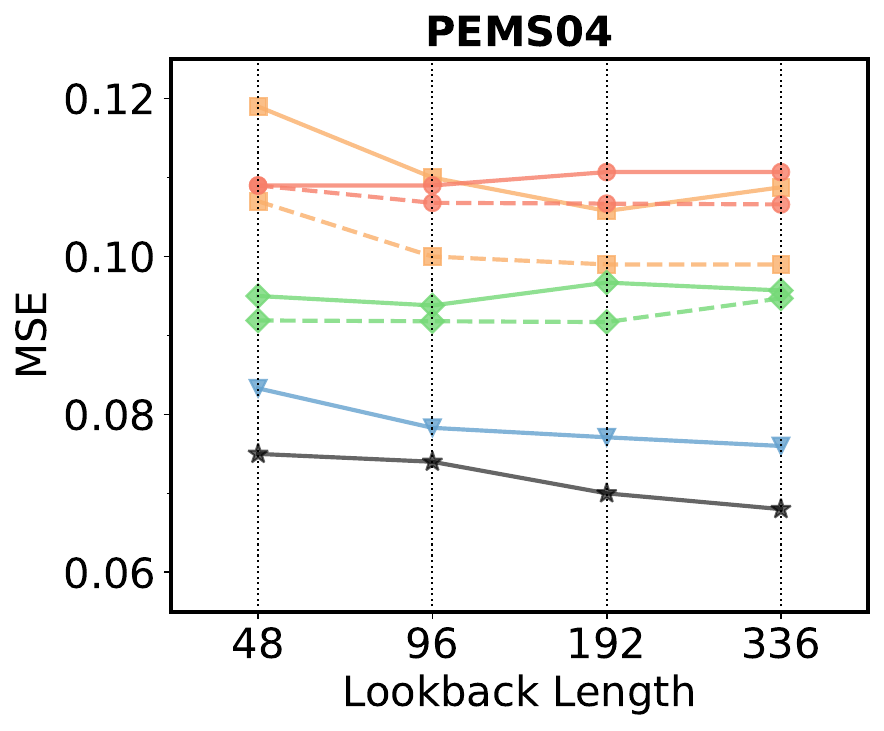}}
    
    \caption{ 
         \protect\includegraphics[height=2.9em]{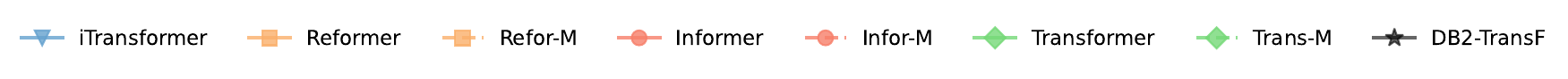}
         \newline
        Ablation results across different datasets using varying input sequence lengths. 
       }
    \label{Ablation_Lookback_Length}
\end{figure*}
Figure~\ref{Ablation_Lookback_Length} illustrates the performance comparison between DB2-TransF and other models as the input lookback length increases.
\begin{enumerate}
    \item Similar to iTransformer, the performance of DB2-TransF improves with an increase in the input lookback length. This performance gain is attributed to learnable Daubechies Wavelet coefficients that efficiently learns the representative features for TSF. 
\item The DB2-TransF model consistently outperforms iTransformer and other Transformer-based variants across all three datasets. Notably, it achieves superior performance compared to the iTransformer model on the Electricity, ETTm1, and PEMS04 datasets.
\item DB2-TransF consistently outperforms most existing models as the input lookback length increases across all three datasets, indicating its ability to effectively capture long-range temporal dependencies while preserving linear computational complexity. 
\end{enumerate}
\subsection{\textbf{Impact of Multi-Scale Levels}}
We conducted an ablation study to investigate the effect of the multi-scale level parameter $L$ on the performance of DB2-TransF for the time series forecasting (TSF) task.

\begin{table*}[htbp!]
\caption{Performance of DB2-TransF for Different Scale Levels}
\label{Scale_Impact}
\centering
\scalebox{0.70}{
\begin{tabular}{|c|c|cc|cc|cc|}
\hline
\multirow{2}{*}{\textbf{Metric}}                                & \multicolumn{1}{l|}{\textbf{Levels}}                             & \multicolumn{2}{c|}{\textbf{1}}                      & \multicolumn{2}{c|}{\textbf{4}}                      & \multicolumn{2}{c|}{\textbf{6}}                      \\ \cline{2-8} 
                                                                & \textbf{\begin{tabular}[c]{@{}c@{}}Input\\ Lengths\end{tabular}} & \multicolumn{1}{c|}{\textbf{MSE}}   & \textbf{MAE}   & \multicolumn{1}{c|}{\textbf{MSE}}   & \textbf{MAE}   & \multicolumn{1}{c|}{\textbf{MSE}}   & \textbf{MAE}   \\ \hline
\multirow{5}{*}{\rotatebox[origin=c]{90}{\textbf{Electricity}}} & \textbf{48}                                                      & \multicolumn{1}{c|}{0.179}          & 0.267          & \multicolumn{1}{c|}{0.176}          & 0.268          & \multicolumn{1}{c|}{0.167}          & 0.260          \\ \cline{2-8} 
                                                                & \textbf{96}                                                      & \multicolumn{1}{c|}{0.148}          & 0.243          & \multicolumn{1}{c|}{0.148}          & 0.248          & \multicolumn{1}{c|}{0.140}          & 0.236          \\ \cline{2-8} 
                                                                & \textbf{192}                                                     & \multicolumn{1}{c|}{0.130}          & 0.224          & \multicolumn{1}{c|}{0.145}          & 0.246          & \multicolumn{1}{c|}{0.133}          & 0.230          \\ \cline{2-8} 
                                                                & \textbf{336}                                                     & \multicolumn{1}{c|}{0.128}          & 0.223          & \multicolumn{1}{c|}{0.138}          & 0.237          & \multicolumn{1}{c|}{0.132}          & 0.230          \\ \cline{2-8} 
                                                                & \textbf{Avg}                                                     & \multicolumn{1}{c|}{\textbf{0.146}} & \textbf{0.239} & \multicolumn{1}{c|}{\textbf{0.152}} & \textbf{0.250} & \multicolumn{1}{c|}{\textbf{0.143}} & \textbf{0.239} \\ \hline
\multirow{5}{*}{\rotatebox[origin=c]{90}{\textbf{ETTm1}}}       & \textbf{48}                                                      & \multicolumn{1}{c|}{0.400}          & 0.395          & \multicolumn{1}{c|}{0.451}          & 0.418          & \multicolumn{1}{c|}{0.432}          & 0.411          \\ \cline{2-8} 
                                                                & \textbf{96}                                                      & \multicolumn{1}{c|}{0.327}          & 0.363          & \multicolumn{1}{c|}{0.333}          & 0.371          & \multicolumn{1}{c|}{0.339}          & 0.376          \\ \cline{2-8} 
                                                                & \textbf{192}                                                     & \multicolumn{1}{c|}{0.325}          & 0.367          & \multicolumn{1}{c|}{0.319}          & 0.364          & \multicolumn{1}{c|}{0.324}          & 0.367          \\ \cline{2-8} 
                                                                & \textbf{336}                                                     & \multicolumn{1}{c|}{0.332}          & 0.373          & \multicolumn{1}{c|}{0.324}          & 0.369          & \multicolumn{1}{c|}{0.322}          & 0.367          \\ \cline{2-8} 
                                                                & \textbf{Avg}                                                     & \multicolumn{1}{c|}{\textbf{0.346}} & \textbf{0.375} & \multicolumn{1}{c|}{\textbf{0.357}} & \textbf{0.381} & \multicolumn{1}{c|}{\textbf{0.354}} & \textbf{0.380} \\ \hline
\multirow{5}{*}{\rotatebox[origin=c]{90}{\textbf{PEMS04}}}      & \textbf{48}                                                      & \multicolumn{1}{c|}{0.075}          & 0.182          & \multicolumn{1}{c|}{0.096}          & 0.205          & \multicolumn{1}{c|}{0.181}          & 0.291          \\ \cline{2-8} 
                                                                & \textbf{96}                                                      & \multicolumn{1}{c|}{0.074}          & 0.182          & \multicolumn{1}{c|}{0.084}          & 0.193          & \multicolumn{1}{c|}{0.076}          & 0.184          \\ \cline{2-8} 
                                                                & \textbf{192}                                                     & \multicolumn{1}{c|}{0.086}          & 0.200          & \multicolumn{1}{c|}{0.070}          & 0.175          & \multicolumn{1}{c|}{0.074}          & 0.180          \\ \cline{2-8} 
                                                                & \textbf{336}                                                     & \multicolumn{1}{c|}{0.077}          & 0.184          & \multicolumn{1}{c|}{0.069}          & 0.173          & \multicolumn{1}{c|}{0.253}          & 0.358          \\ \cline{2-8} 
                                                                & \textbf{Avg}                                                     & \multicolumn{1}{c|}{\textbf{0.078}} & \textbf{0.187} & \multicolumn{1}{c|}{\textbf{0.080}} & \textbf{0.187} & \multicolumn{1}{c|}{\textbf{0.146}} & \textbf{0.253} \\ \hline
\end{tabular}}
\end{table*}

 The study was carried out on three benchmark datasets: Electricity, PEMS04, and ETTm1. The optimal performance on the Electricity and PEMS04 datasets was achieved when $L = 1$, whereas for the ETTm1 dataset, the best results were obtained with $L = 6$. These findings indicate that the optimal choice of $L$ varies across datasets. Specifically, increasing $L$ led to a decline in performance for the Electricity and PEMS04 datasets, while for ETTm1, performance improved as $L$ increased.

\section{Analysis of Computational Complexity}

\begin{table}[htbp!]
\caption{Memory Overhead and Training Time Per Epoch for DB2-TransF}
\label{Computational_Complexity}
\centering 
\scalebox{0.70}{
\begin{tabular}{|ccc|ccc|ccc|c|}
\hline
\multicolumn{3}{|c|}{\textbf{Level 1}}                                                               & \multicolumn{3}{c|}{\textbf{Level 4}}                                                               & \multicolumn{3}{c|}{\textbf{Level 6}}                                                               & \textbf{Dataset}                 \\ \hline
\multicolumn{1}{|c|}{\textbf{Length}} & \multicolumn{1}{c|}{\textbf{GPU Usage}} & \textbf{Time/Iteration} & \multicolumn{1}{c|}{\textbf{Length}} & \multicolumn{1}{c|}{\textbf{GPU Usage}} & \textbf{Time/Iteration} & \multicolumn{1}{c|}{\textbf{Length}} & \multicolumn{1}{c|}{\textbf{GPU Usage}} &\textbf{Time/Iteration} & \multirow{5}{*}{\textbf{\rotatebox[origin=c]{90}{\textbf{Electricity}}}}    \\ \cline{1-9}
\multicolumn{1}{|c|}{48}           & \multicolumn{1}{c|}{0.405GB}        & 195ms            & \multicolumn{1}{c|}{48}           & \multicolumn{1}{c|}{0.494GB}        & 534ms            & \multicolumn{1}{c|}{48}           & \multicolumn{1}{c|}{0.508 GB}       & 784ms            &                                  \\ \cline{1-9}
\multicolumn{1}{|c|}{96}           & \multicolumn{1}{c|}{0.407GB}        & 192ms            & \multicolumn{1}{c|}{96}           & \multicolumn{1}{c|}{0.496GB}        & 536ms            & \multicolumn{1}{c|}{96}           & \multicolumn{1}{c|}{0.510 GB}       & 759ms            &                                  \\ \cline{1-9}
\multicolumn{1}{|c|}{192}          & \multicolumn{1}{c|}{0.413GB}        & 190ms            & \multicolumn{1}{c|}{192}          & \multicolumn{1}{c|}{0.515GB}        & 510ms            & \multicolumn{1}{c|}{192}          & \multicolumn{1}{c|}{0.517 GB}       & 740ms            &                                  \\ \cline{1-9}
\multicolumn{1}{|c|}{336}          & \multicolumn{1}{c|}{0.421GB}        & 187ms            & \multicolumn{1}{c|}{336}          & \multicolumn{1}{c|}{0.508GB}        & 558ms            & \multicolumn{1}{c|}{336}          & \multicolumn{1}{c|}{0.523GB}        & 729ms            &                                  \\ \hline
\multicolumn{1}{|c|}{48}           & \multicolumn{1}{c|}{0.037GB}        & 116ms            & \multicolumn{1}{c|}{48}           & \multicolumn{1}{c|}{0.038GB}        & 300ms            & \multicolumn{1}{c|}{48}           & \multicolumn{1}{c|}{0.039GB}        & 395ms            & \multirow{4}{*}{\textbf{\rotatebox[origin=c]{90}{\textbf{ETTm1}}}}  \\ \cline{1-9}
\multicolumn{1}{|c|}{96}           & \multicolumn{1}{c|}{0.038GB}        & 122ms            & \multicolumn{1}{c|}{96}           & \multicolumn{1}{c|}{0.039GB}        & 304ms            & \multicolumn{1}{c|}{96}           & \multicolumn{1}{c|}{0.039GB}        & 404ms            &                                  \\ \cline{1-9}
\multicolumn{1}{|c|}{192}          & \multicolumn{1}{c|}{0.038GB}        & 103ms            & \multicolumn{1}{c|}{192}          & \multicolumn{1}{c|}{0.039GB}        & 291ms            & \multicolumn{1}{c|}{192}          & \multicolumn{1}{c|}{0.040GB}        & 383ms            &                                  \\ \cline{1-9}
\multicolumn{1}{|c|}{336}          & \multicolumn{1}{c|}{0.039GB}        & 105ms            & \multicolumn{1}{c|}{336}          & \multicolumn{1}{c|}{0.040GB}        & 287ms            & \multicolumn{1}{c|}{336}          & \multicolumn{1}{c|}{0.041GB}        & 390ms            &                                  \\ \hline
\multicolumn{1}{|c|}{48}           & \multicolumn{1}{c|}{3.580GB}        & 940ms            & \multicolumn{1}{c|}{48}           & \multicolumn{1}{c|}{4.312GB}        & 1755ms           & \multicolumn{1}{c|}{48}           & \multicolumn{1}{c|}{4.445GB}        & 2473             & \multirow{4}{*}{\textbf{\rotatebox[origin=c]{90}{\textbf{PEMS04}}}} \\ \cline{1-9}
\multicolumn{1}{|c|}{96}           & \multicolumn{1}{c|}{3.585GB}        & 937ms            & \multicolumn{1}{c|}{96}           & \multicolumn{1}{c|}{4.354GB}        & 1788ms           & \multicolumn{1}{c|}{96}           & \multicolumn{1}{c|}{4.420GB}        & 2567             &                                  \\ \cline{1-9}
\multicolumn{1}{|c|}{192}          & \multicolumn{1}{c|}{3.591GB}        & 932ms            & \multicolumn{1}{c|}{192}          & \multicolumn{1}{c|}{4.361GB}        & 1789ms           & \multicolumn{1}{c|}{192}          & \multicolumn{1}{c|}{4.427GB}        & 2553             &                                  \\ \cline{1-9}
\multicolumn{1}{|c|}{336}          & \multicolumn{1}{c|}{3.596GB}        & 927ms            & \multicolumn{1}{c|}{336}          & \multicolumn{1}{c|}{4.370GB}        & 1740ms           & \multicolumn{1}{c|}{336}          & \multicolumn{1}{c|}{4.436GB}        & 2534             &                                  \\ \hline
\end{tabular}}
\end{table}
The following key observations can be drawn from Table~\ref{Computational_Complexity}, which reports the memory overhead and training time per epoch for DB2-TransF at different decomposition levels across the Electricity, ETTm1, and PEMS04 datasets.

\begin{itemize}
    \item \textbf{Higher wavelet decomposition levels lead to increased computational cost:} For all datasets (Electricity, ETTm1, and PEMS04), moving from Level 1 to Level 4 and Level 6 results in greater GPU memory usage and longer training times per epoch. For instance, in the PEMS04 dataset with input length 48, GPU consumption and training time rise from 3.580GB and 940ms at Level 1 to 4.445GB and 2473ms at Level 6.

    \item \textbf{Input sequence length has a limited impact compared to decomposition level:} Varying the input length (from 48 to 336) within the same wavelet level yields only minor changes in GPU usage and training time. For example, in the Electricity dataset at Level 6, GPU usage increases marginally from 0.508GB to 0.523GB, while training time slightly decreases from 784ms to 729ms.

    \item \textbf{Dataset complexity significantly influences resource requirements:} The memory and time overhead differ substantially between datasets. Among the three, PEMS04 demands the most resources, while ETTm1 is the most lightweight. At Level 6 with input length 48, PEMS04 requires 4.445GB and 2473ms, whereas ETTm1 needs only 0.039GB and 395ms.

    \item \textbf{Training time does not always scale linearly with input length:} In certain cases, increasing the input sequence length results in a slight reduction in training time. This non-linear behavior may stem from GPU scheduling, memory bandwidth optimization, or internal batching mechanisms. For instance, in the Electricity dataset at Level 6, training time decreases from 784ms (length 48) to 729ms (length 336).
\end{itemize}

\section{Conclusion and Future Work}
In this paper, we present a lightweight deep model that overcomes the computational overhead problem of the transformer-based deep model that uses a self-attention mechanism that causes quadratic complexity. To overcome the limitation mentioned above, we crafted a novel MLDB block that employs learnable and multi-scale Daubechies wavelet layers for extracting approximate and detailed information for the time series task while ensuring the effective modeling of inter-variable correlations. The efficacy of the proposed model was evaluated across 13 time series forecasting datasets. The experimental results illustrate that the proposed TSF model outperformed the existing state-of-the-art transformer-based models and also requires significantly less computational overhead compared to the existing methods. The proposed DB2-TransF can act as a better alternative to the self-attention-based transformer models. In the future, we would like to explore and evaluate the performance of the proposed DB2-TransF in other domains such as computer vision, natural language processing, and audio classification.

 \bibliographystyle{elsarticle-num} 
 \bibliography{DB2_Trans}





\end{document}